%% file: main.tex
\documentclass{article}
\usepackage[preprint]{neurips_2025}
\usepackage{fancyhdr}
\usepackage{amsmath,amsfonts,bm,amsthm}
\usepackage{algorithm}
\usepackage{algpseudocode}
\newtheorem{definition}{Definition}
\usepackage{makecell}
\usepackage{subcaption}
\usepackage[most]{tcolorbox}
\usepackage{minted}
\usepackage{courier} 
\usepackage{enumitem}
\usepackage{booktabs}
\usepackage{hyperref} 
\usepackage{color, colortbl, xcolor}
\usepackage{pifont, dsfont}
\DeclareMathOperator*{\concat}{%
    \mathchoice%
        {\Big\Vert}%
        {\big\Vert}%
        {\Vert}%
        {\Vert}%
}
\newtheorem{theorem}{Theorem}

\title{TransactionGPT}

\author{
\\
Visa Research
}

\AtBeginDocument{\pagestyle{fancy} \fancyhf{} \fancyhead[C]{TransactionGPT} \fancyhead[R]{Visa Research} \fancyfoot[C]{\thepage} }

\begin{document}

\maketitle


\begin{abstract}
We present \textbf{TransactionGPT} (TGPT), a foundation model for consumer transaction data within one of the world's largest payment networks.
TGPT is designed to understand and generate transaction trajectories while simultaneously supporting a variety of downstream prediction and classification tasks.
We introduce a novel 3D-Transformer architecture specifically tailored for capturing the complex dynamics in payment transaction data.
This architecture incorporates design innovations that enhance modality fusion and computational efficiency, while seamlessly enabling joint optimization with downstream objectives.
Trained on billion‑scale real‑world transactions, TGPT significantly improves downstream anomaly transaction detection performance against a competitive production model and exhibits advantages over baselines in generating future transactions. 
We conduct extensive empirical evaluations utilizing a diverse collection of company transaction datasets spanning multiple downstream tasks, thereby enabling a thorough assessment of TGPT’s effectiveness and efficiency in comparison to established methodologies. Furthermore, we examine the incorporation of LLM-derived embeddings within TGPT and benchmark its performance against fine-tuned LLMs, demonstrating that TGPT achieves superior predictive accuracy as well as faster training and inference.
We anticipate that the architectural innovations and practical guidelines from this work will advance foundation models for transaction-like data and catalyze future research in this emerging field.
\end{abstract}

\input{sec01-intro}
\input{sec02-prelim}

\input{sec03-method}

\input{sec04-recipe}
\input{sec05-exp}
\input{sec06-discuss}
\input{sec07-rw}

\section{Conclusion and Future Work}
This work presents TransactionGPT (TGPT), a foundation model that captures complex consumer shopping dynamics from Multi-Modal-Temporal-Tabular (MMTT) data.
Our novel 3D-Transformer architecture, augmented by a virtual token mechanism, is designed to effectively encode and fuse diverse data modalities.
Extensive experiments on large-scale, real-world payment data validate TGPT's ability to learn meaningful transaction patterns, leading to significant performance improvements on critical downstream tasks.
Furthermore, we quantify the benefits of several designs that enhance the TGPT's efficiency and scalability.
This research opens a new avenue for foundation modeling of ubiquitous MMTT data on web.
Future directions include accelerating model performance, developing superior multi-modal encoders, and exploring the joint optimization of MMTT foundation models with LLMs.

\newpage

\textbf{Core Contributors}

Yingtong Dou, Zhimeng Jiang, Tianyi Zhang, Mingzhi Hu, Zhichao Xu, Yuzhong Chen

\vspace{0.5cm}

\textbf{Contributors}

Shubham Jain, Uday Singh Saini, Xiran Fan, Jiarui Sun, Menghai Pan,
Junpeng Wang, Xin Dai, Liang Wang, Chin-Chia Michael Yeh, Yujie Fan, Yan Zheng, Vineeth Rakesh, Huiyuan Chen, Guanchu Wang, Mangesh Bendre, Zhongfang Zhuang, Xiaoting Li, Prince Aboagye, Vivian Lai, Minghua Xu, Hao Yang, Yiwei Cai, Mahashweta Das

\vspace{0.5cm}
\textbf{Project Lead}

Yuzhong Chen

\newpage
\bibliographystyle{abbrvnat}
\bibliography{sample-base}

\newpage
\appendix
\input{appendix}

\end{document}

%% file: sec01-intro.tex
\section{Introduction}
Foundation models have emerged as a transformative paradigm in artificial intelligence, characterized by their ability to acquire broad capabilities from self-supervised training on large-scale data and to adapt effectively to a wide range of downstream tasks~\citep{bommasani2021opportunities}.
By encoding rich prior knowledge, these models serve as a common backbone across domains from natural language processing~\citep{radford2019language} to computer vision~\citep{dosovitskiy2020image} and even specialized fields such as the medical industry~\cite{moor2023foundation}.
A prominent and highly successful class of foundation models is built upon the Transformer architecture~\citep{vaswani2017attention}, whose self-attention mechanism excels at modeling intricate dependencies between tokens in sequential and structured data.
Scaling Transformer-based models to billions of parameters in accordance with empirical scaling laws~\citep{kaplan2020scaling} has yielded powerful systems that now underpin state-of-the-art language and vision applications.

\begin{figure}
    \centering
    \includegraphics[width=0.9\linewidth]{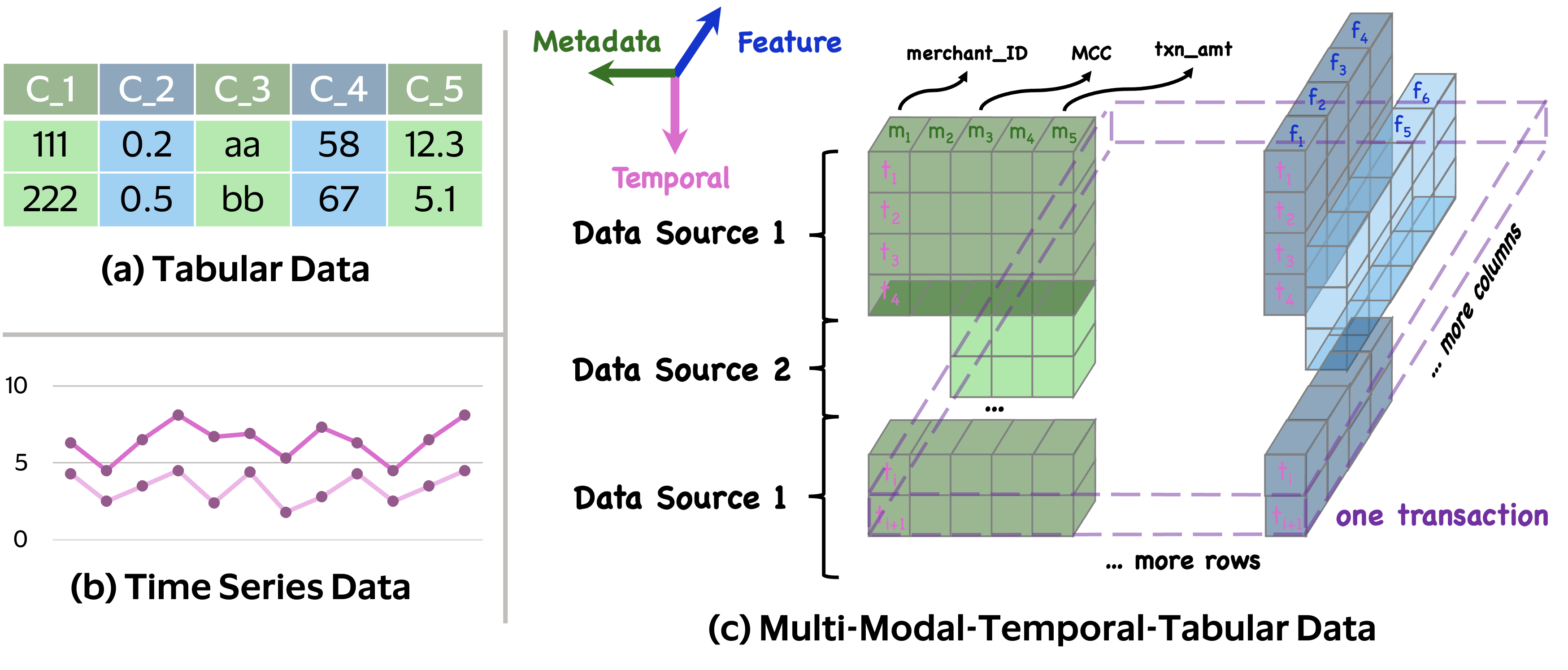}
    \caption{A comparison of multi-modal temporal tabular payments data and other similar data types.}
    \label{fig:mmtt_data}
\end{figure}


Despite the widespread adoption of foundation models, only a few endeavors~\citep{skalski2023towards, unbox2025behaviorgpt, li2025panther, aminian2025fraudtransformer} have explored their application to payment data, and others only disclose vague or incomplete information~\citep{stripe2025news}.
While limited access to real-world payment network data may contribute to this gap, the primary obstacle is the large-scale and complex structural nature of \textbf{M}ulti-\textbf{M}odal-\textbf{T}emporal-\textbf{T}abular (\textbf{MMTT}) transaction data.
As shown in Fig.~\ref{fig:mmtt_data}, a temporal transaction sequence from the same consumer often comprises transactions from multiple sources; meanwhile, each transaction has diverse multi-modality fields including metadata and downstream features (more details in Sec.~\ref{sec:prelim-data}).
Transformer-based architectures provide a natural framework for encoding transaction sequence data.
Just as language models predict future words based on preceding context, we posit that consumer transaction trajectories can be forecast using historical transactions.


We propose to learn invariant patterns in individual and collective shopping behavior through foundation model training on large-scale transaction data.
However, applying existing Transformer architectures to encode transaction sequences is non-trivial. 
\textbf{First}, payment transaction data possesses domain-specific characteristics, with predominantly numerical and categorical fields that have limited semantic richness~\citep{bahnsen2016feature}.
This makes direct LLM adoption both computationally costly and ineffective, as their linguistic capabilities provide minimal value for transaction data comprehension.
\textbf{Second}, transaction data are stored as database tables. Existing tabular foundation models~\citep{hollmann2025accurate,hegselmann2023tabllm}, however, are inadequate for this domain as they either handle only small-scale data or require semantically rich column names.
Furthermore, most are tailored for classification tasks and lack the flexibility to support the varied downstream applications essential for industrial payment systems.
\textbf{Third}, although transaction sequences bear superficial resemblance to time series~\citep{wen2023transformers} or point process~\citep{mei2017neural} data, they differ fundamentally in data complexity: time series or point process consist of single or multiple scalar values per timestep, while transactions contain multi- and high-dimensional information, including metadata with various cardinalities and derived features.
This structural difference makes existing temporal data foundation models~\citep{garza2023timegpt, ekambaram2024tiny} inadequate for transaction sequence modeling.
We argue that a foundation model for the consumer payment industry should accommodate: 
1) the complex, heterogeneous nature of MMTT payment data;
2) diverse requirements across downstream applications;
3) stringent efficiency and latency constraints.


To overcome these challenges and develop a foundation model tailored to the consumer payment industry, this paper presents \textbf{T}ransaction \textbf{G}enerative \textbf{P}redictive \textbf{T}ransformer (\textbf{TransactionGPT} or \textbf{TGPT}), 
a specialized foundation model for transaction-like MMTT data, with direct relevance and broad applicability for the consumer payment industry.
TransactionGPT is a Transformer-based model trained on billions of transactions to address various downstream tasks in payment, including prediction, classification, and representation learning.
The model features a novel 3-Dimensional (3D) architecture that utilizes three specialized Transformers to hierarchically encode transaction features, individual transaction representations, and sequential transaction patterns.
Moreover, we introduce a novel \textit{virtual token mechanism} that enables effective modality fusion between different architectural dimensions. This design dramatically reduces the computational complexity of the 3D-Transformer structure while boosting performance across downstream applications. Notably, the TransactionGPT architecture maintains high flexibility through modular, switchable components at each dimension and layer, facilitating the integration of emerging Transformer and generative modeling innovations.




We highlight our contributions as follows:
\begin{itemize}
    \item We introduce TransactionGPT, a foundation model for understanding and generating consumer transaction trajectories, trained on billion-scale data from one of the world's largest payment networks. It includes a 3D-Transformer for encoding complex transaction data and a virtual token mechanism to enable effective cross-modality information fusion. 
    \item We present a series of techniques developed to ensure the scalability and efficiency of TransactionGPT for real-world deployment and quantify their performance impact. We also provide a detailed analysis of our architectural design choices, offering insights and lessons learned from the model's development and evaluation to benefit the practitioners in the community. 
    \item We evaluate TGPT under three payment-related tasks to demonstrate its foundational capability in transaction anomaly detection, generation, and representation learning. The results show that TransactionGPT achieves a 22\% relative improvement over the production model on a business-critical anomaly detection metric and demonstrates superior efficiency compared to LLM-based foundation models. 
    
\end{itemize}

%% file: sec02-prelim.tex
\section{Preliminary}
In this section, we characterize the complex and hierarchical structure of transaction data, establish the notations used throughout the paper, and formally define our problem.

\subsection{Multi-Modal-Temporal-Tabular Data}
\label{sec:prelim-data}

\textbf{Temporal-tabular structure}.
The transaction sequence from a single payment account is given in a temporal-tabular format, denoted by $\mathcal{S} = [\mathbf{tr}_1, \mathbf{tr}_2, \cdots, \mathbf{tr}_n]$, where $\mathbf{tr}_1$ is the earliest transaction and $\mathbf{tr}_n$ is the latest. Each transaction $\mathbf{tr}$ is extracted from the transaction table and represented as a $d$-dimensional vector comprising the following components in tabular form:
\begin{equation}
    \mathbf{tr} = [\mathcal{M} \oplus \mathcal{E} \oplus \mathcal{F}]  \in \mathbb{R}^{d},
\label{eq:txn_vector}
\end{equation}
where $\mathcal{M}$ denotes the vector of metadata associated with $\mathbf{tr}$, including essential numerical attributes such as the transaction amount and timestamp.
$\mathcal{E}$ represents the one-hot entity vectors linked to the current transaction.
Common entities include merchant ID and merchant category.
Entities are part of metadata in a broader sense; we use a distinct symbol $\mathcal{E}$ because they are encoded differently from other categorical metadata (see Sec.~\ref{sec:method-txn-encoder}).
Meanwhile, the embedding of $\mathcal{E}$ can be dynamically obtained as the output of another model or precomputed and retrieved from a vector database.
Metadata $\mathcal{M}$ and entities $\mathcal{E}$ are present in every transaction and are independent of any downstream task.
Unless otherwise specified, the term \textit{metadata} in this paper refers to $\mathcal{M} \oplus \mathcal{E}$.
Additionally, $\mathcal{F}$ is an optional feature vector based on the availability of downstream task-dependent features.
$\mathcal{F}$ usually consists of hundreds to thousands of numerical and categorical values, depending on the specific requirements of the downstream task.

\noindent \textbf{Multi-modal structure}.
The metadata fields $\mathcal{M} \oplus \mathcal{E}$ (typically totaling 10 to 20 fields, mostly categorical) within the same $\mathbf{tr}$ collectively describe the basic property of $\mathbf{tr}$ and are highly correlated with each other.
For example, the merchant category, merchant name, location, and price are all interdependent when purchasing a pair of shoes.
Moreover, some metadata fields exhibit an extremely high cardinality, e.g., merchant ID may have millions of unique values, necessitating large embedding sizes during data encoding.
In contrast, the feature vector $\mathcal{F}$ does not contain explicit transaction metadata but includes features specifically crafted for a downstream task.
$\mathcal{F}$ often consists of hundreds or thousands of scalar values (mostly numerical), each with low cardinality and thus small embedding sizes.
Accordingly, metadata and feature vectors effectively constitute \textit{two distinct data modalities}, which are embedded and modeled using different strategies within TGPT.
Consequently, transaction data exhibits a multi-modal-temporal-tabular (\textbf{MMTT}) structure in general, as shown in Fig.~\ref{fig:mmtt_data}.

\subsection{Problem Definition}
\label{sec:prelim-problem}

Analogous to how LLMs are pretrained to generate future text tokens~\citep{radford2019language, touvron2023llama}, TGPT is trained on MMTT transaction data to generate future transaction trajectories and predict transaction properties.
We formally define the problem as follows:
\begin{definition}
Given a set of MMTT transaction sequences $\{\mathcal{S}\}, \mathcal{S} \in \mathcal{S}_{train}$, our goal is to train a Transformer-based model $f$ (i.e., TGPT) to encode the transaction information and learn the temporal dynamics at scale.
Given the historical transaction sequences $\mathcal{S}_{test} = [\mathbf{tr}_1, \mathbf{tr}_2, \cdots, \mathbf{tr}_n]$, an optimal TransactionGPT should be able to generate the next $t$ transactions
$[\mathbf{tr}_{n+1}, \mathbf{tr}_{n+2}, \ldots, \mathbf{tr}_{n+t}]$.
For each generated transaction $\mathbf{tr}_{n+i} (1\leq i \leq t)$, the model can also produce a predicted label $\mathbf{L}_{n+i}$ if required by a downstream classification task.
\end{definition}

With the data and problem formally defined, we now turn to the design of a modeling framework capable of effectively capturing the complex characteristics of MMTT transaction data.
The following section first summarizes the architectural evolution of TGPT and presents the technical details of each model component.


%% file: sec03-method.tex
\section{Methodology}
\label{sec:method}

\begin{figure*}
    \centering
    \includegraphics[width=1.0\linewidth]{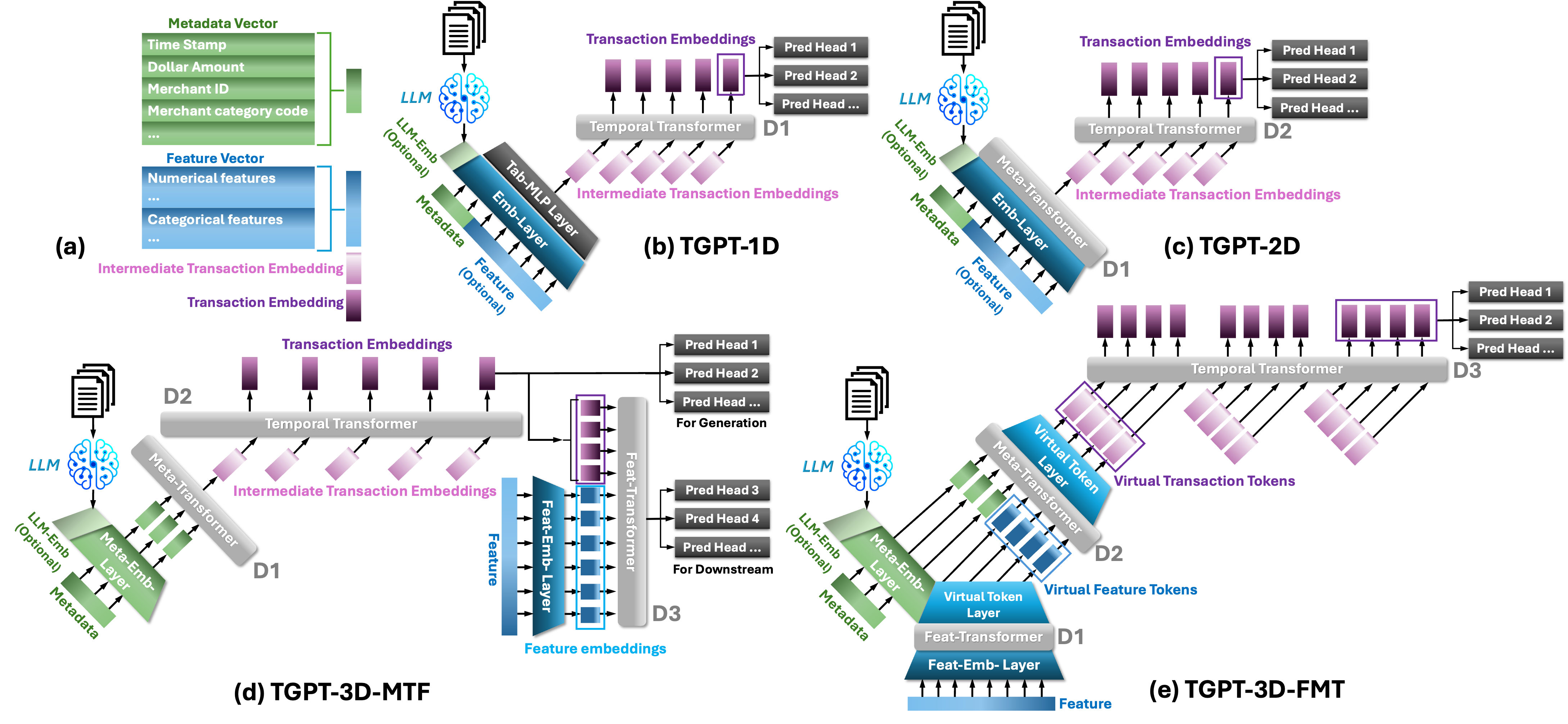}
    \caption{TransactionGPT (TGPT) model architecture evolution. (a)  Diagram of the metadata vector, feature vector, intermediate transaction embedding, and final transaction embedding. (b) TGPT-1D. (c) TGPT-2D. (d) TGPT-3D-\textbf{MTF} (\textbf{M}etadata-\textbf{T}emporal-\textbf{F}eature). (e) TGPT-3D-\textbf{FMT} (\textbf{F}eature-\textbf{M}etadata-\textbf{T}emporal).}
    \label{fig:main_figure}
\end{figure*}


In this section, we first summarize the evolution path of the TGPT model from 1- to 3-Dimensional and innovations in the fusion of modality between Transformers in different dimensions (Sec.~\ref{sec:method-evolution}).
Then we introduce the technical details of Transformer in each dimension (Sec.~\ref{sec:method-transformer}) and our novel modality fusion method (Sec.~\ref{sec:modality_fusion}), 
followed by model time complexity analysis (Sec.~\ref{sec:complexity}) and optimization (Sec.~\ref{sec:loss_function}).

\subsection{Model Architecture Evolution}
\label{sec:method-evolution}
During the process of acquiring deeper insights into MMTT data and designing more effective model components to capture the intrinsic patterns of consumer payment trajectories, the architecture of TransactionGPT has undergone a gradual evolution.
Initially, we employ a straightforward single temporal Transformer (1D), which is subsequently advanced to a structure comprising two inter-coupled Transformers for metadata and temporal modality encoding (2D), and ultimately to three inter-coupled transformers (3D) with an additional Transformer to encode features.
Each stage introduces necessary sophistication to better model the underlying data characteristics, as shown in Fig.~\ref{fig:main_figure}.
Each additional Transformer—corresponding to an added dimension (data modality) in the architecture—is introduced to more effectively encode information from its respective data modality.

\noindent \textbf{TGPT-1D.} 
Only a single Transformer is used to model the temporal dependencies between transactions along the payment trajectory $\mathcal{S}$, as shown in Fig.~\ref{fig:main_figure}(b).
Meanwhile, the interaction among the various fields within the same transaction is captured by a tabular MLP block.
Sec.~\ref{sec:method-seq-encoder} has more details about the temporal Transformer $\mathsf{TF}(\mathcal{S})$.

\noindent \textbf{TGPT-2D.}
To more effectively encode the interactions between different data fields within one transaction $\mathbf{tr}$ (Eq.~\ref{eq:txn_vector}), an additional Transformer is incorporated, replacing the tabular MLP, resulting in a 2D transformer architecture, as shown in Fig.~\ref{fig:main_figure}(c).
In Sec.~\ref{sec:method-txn-encoder}, we present how we process and encode different transaction metadata $\mathcal{M}$ to facilitate their interaction through the metadata Transformer $\mathsf{TF}(\mathbf{tr})$.

\noindent \textbf{TGPT-3D.}
As we adopt TGPT towards downstream tasks with domain specific features, we think a separate feature transformer $\mathsf{TF}(\mathcal{F})$ is necessary.
The reason has two folds:
\textbf{First}, the feature set $\mathcal{F}$ is way larger than the metadata and entity set ($|\mathcal{F}| \gg |\mathcal{M}| + |\mathcal{E}|$); features within $\mathcal{F}$ are highly correlated, but their semantics are only related to the corresponding downstream task ~\citep{bahnsen2016feature}.
\textbf{Second}, metadata and downstream features have conflicting embedding requirements: high-cardinality fields like merchant ID with millions of unique values need large embedding sizes to capture their rich information, while downstream features can be effectively represented with much smaller embeddings.
In the 2D Transformer, all fields must share the same embedding size, creating an impossible trade-off where large embeddings cause excessive computational costs and overfitting risk for downstream features, while small embeddings fail to provide sufficient expressive capacity for high-cardinality metadata fields.
Therefore, encoding features and metadata with distinct Transformers allows each to use optimally-sized embeddings tailored to their specific requirements, resulting in a 3D Transformer architecture, as shown in Fig.~\ref{fig:main_figure}(d) and (e). 
Sec.~\ref{sec:method-feature-encoder} has extended details about the feature Transformer $\mathsf{TF}(\mathcal{F})$.

\noindent \textbf{Modality Fusion.}
Once the three Transformers are designated to encode and generate embeddings for the temporal, metadata, and feature modalities, the challenge of designing an effective modality embedding fusion mechanism becomes both critical and complex.
Specifically, the small-sized embedding output of $\mathsf{TF}(\mathcal{F})$ need to be merged with the large-sized embeddings of metadata fields into a unified representation that encapsulates all relevant information for a single transaction $\mathbf{tr}$, which can be efficiently processed by $\mathsf{TF}(\mathbf{tr})$ and $\mathsf{TF}(\mathcal{S})$.
A naive approach, concatenating all feature and metadata embeddings into a single vector and feeding it directly into $\mathsf{TF}(\mathcal{S})$, is costly in terms of parameter count and processing time.
Extensive experiments also reveal that employing an MLP to rescale the concatenated embeddings to a manageable size for $\mathsf{TF}(\mathcal{S})$ leads to suboptimal performance (see Sec.~\ref{sec:exp-cls}).
The above compression approach introduces an information bottleneck: the small embedding size restricts information flow, while embeddings large enough for adequate information preservation dramatically increase computational costs.
To address the challenge, we propose a novel \textbf{virtual token} based 2-step modality fusion mechanism.
This approach first performs feature-to-transaction modality fusion via \textbf{virtual feature tokens}, followed by transaction-to-temporal modality fusion via \textbf{virtual transaction tokens}.
The details of this mechanism are presented in Sec.~\ref{sec:virtual_token}. 

\noindent \textbf{Transaction Embedding Formulations.} The architectural differences among the 1D, 2D, and 3D model variants lead to distinct computational formulations of the transaction-level embedding, summarized in Table~\ref{tab:trans-embed} and detailed in Sections~\ref{sec:method-transformer} and~\ref{sec:modality_fusion}.

\begin{table*}[t]
    \centering
    \caption{The computation formulations of the embedding of a transaction $\mathbf{tr}$ given $\mathcal{M}, \mathcal{E}, \mathcal{F} \in \mathbf{tr}$. Operation $\mathsf{Integrate}$ aggregates a 2D tensor (a set of embedding vectors) into a 1D tensor (a single embedding vector). Its design choices are discussed in Sec.~\ref{sec:train_tips}. }
    \centering
    \resizebox{0.999\textwidth}{!}{%
    \begin{tabular}{l|l|l}
        \toprule
        \textbf{Model}  & \textbf{Transaction Embedding Formulation} & \textbf{Structure}\\
        \hline
        {TGPT-1D}  & ${E}^{\mathbf{tr}} = \mathsf{MLP}( \mathrm{Emb}(\mathcal{M})\oplus\mathrm{Emb}(\mathcal{E}) \oplus\mathrm{Emb}(\mathcal{F}))$ &  1D: $1 \times d_\mathbf{tr}$ \\
        {TGPT-2D}   & ${E}^{\mathbf{tr}} =  \mathsf{Integrate}(\mathsf{TFEncoder}(\mathrm{Emb}(\mathcal{M})\oplus\mathrm{Emb}(\mathcal{E}) \oplus\mathrm{Emb}(\mathcal{F})))$, see Eq.~\ref{eq:meta-feat-transformer} & 1D: $1 \times d_\mathbf{tr}$\\
        {TGPT-3D-\textbf{MTF}}  &  ${E}^{\mathbf{tr}} = \mathsf{Integrate}(\mathsf{TFEncoder}(\mathrm{Emb}(\mathcal{M})\oplus\mathrm{Emb}(\mathcal{E})))$, see Eq.~\ref{eq:meta-transformer} & 1D: $1 \times d_\mathbf{tr}$  \\
        {TGPT-3D-\textbf{FMT}} &  $\mathbf{E}^{\mathbf{tr}} = \mathsf{VTL}(\mathsf{TFEncoder}(\mathrm{Emb}(\mathcal{M})\oplus\mathrm{Emb}(\mathcal{E}) \oplus\mathsf{VTL}(\mathsf{TFEncoder}(\mathcal{F}))))$, see Eq.~\ref{eq:virtual_trans_tokens} & 2D: $v_t \times \frac{d_\mathbf{tr}}{v_t}$ \\
        \bottomrule
    \end{tabular}
    }
    \label{tab:trans-embed}
\end{table*}

\subsection{Three Transformers}
\label{sec:method-transformer}

\subsubsection{Temporal Transformer}
\label{sec:method-seq-encoder}

As depicted in Fig.~\ref{fig:main_figure} (a), we feed a transaction sequence $\mathcal{S}$ into the Temporal Transformer $\mathsf{TF}(\mathcal{S})$ after we obtain each transaction's embedding $E^{\mathbf{tr}}\in \mathbb{R}^{1 \times d_\mathbf{tr}}$, where $d_{\mathbf{tr}}$ is the transaction embedding dimension.
We choose the \textit{decoder-only} Transformer~\citep{vaswani2017attention, radford2019language} as the backbone model. 
Specifically, each transaction's embedding is regarded as a token and fed into stacked standard Transformer decoder blocks (abbr. $\mathsf{TFDecoder}$) with multi-head self attention:
\begin{align}
     \mathsf{TF}(\mathcal{S}) & = \mathsf{TFDecoder}(\{{E}^{\mathbf{tr_{1}}}, \cdots, {E}^\mathbf{tr_{n}}\}, \mathbf{tr}\in \mathcal{S}), \\
     {E}^{\mathbf{tr}_{n+1}} & = \mathsf{TF}(\mathcal{S})[\mathbf{tr}_{n}, :].
\label{eq:seq_encoder}
\end{align}
The decoder-only architecture enables TGPT to auto-regressively generate future transactions, where $\mathbf{tr}_1$ to $\mathbf{tr}_{n}$ are the historical transactions encoded by $\mathsf{TF}(\mathcal{S})$,
$\mathbf{tr}_{n+1}$ is next transaction to be generated.  
The output of $\mathsf{TF}(\mathcal{S})$ for last transaction $\mathbf{tr}_{n}$ is the embedding of next transaction ${E}^{\mathbf{tr}_{n+1}}\in \mathbb{R}^{1 \times d_\mathbf{tr}}$.

\noindent\textbf{Local Attention.}
To improve the scalability of $\mathsf{TF}(\mathcal{S})$ over long transaction sequences, we empirically find that the local attention mechanism~\citep{beltagy2020longformer} has a decent balance between input sequence length and downstream task performance.
Specifically, the local attention mechanism restricts each transaction in $\mathcal{S}$ only attends to transactions in its context window with length equal to $w$, reducing the time complexity of causally masked attention from $O(|\mathcal{S}|^2 \times d_{\mathbf{tr}})$ to $O(w^2 \times d_{\mathbf{tr}})$, where $w \ll |\mathcal{S}|$.
Please refer to Sec.~\ref{sec:exp-efficiency} for its performance comparison with full attention.

Note that we also explored the \textit{encoder-only} architecture for $\mathsf{TF}(\mathcal{S})$, and its next one transaction generation performance is similar to $\mathsf{TFDecoder}$.
Extended discussions on Transformer design choices are presented in Sec.~\ref{sec:train_tips}.

\subsubsection{Feature Transformer}
\label{sec:method-feature-encoder}
According to our discussion in Sec.~\ref{sec:method-evolution}, the feature encoder is a separate tabular Transformer~\citep{huang2020tabtransformer} similar to $\mathsf{TF}(\textbf{tr})$ but dedicated to downstream-dependent feature encoding:
\begin{align}
    \mathsf{TF}(\mathcal{F}) & = \mathsf{TFEncoder}(\mathrm{Emb}(\mathcal{F})),
\end{align}
each ${E}_{f} \in \mathbb{R}^{1 \times d_f}, f\in\mathcal{F}$ is obtained by linear projection $\mathrm{Emb}(\mathcal{F})$. 

\subsubsection{Metadata Transformer}
\label{sec:method-txn-encoder}
The Metadata Transformer $\mathsf{TF}(\mathbf{tr})$ employs an encoder-only backbone with bi-directional attention as we need to fully exploit the interactions between different data fields~\citep{huang2020tabtransformer}.
Since the encoder architecture is standard, we emphasize more on the non-trivial part, which is encoding different types of metadata as uni-sized embeddings as Transformer input.
According to Sec.~\ref{sec:prelim-data}, each transaction has two types of metadata $\mathcal{M}$ and $\mathcal{E}$.
Therefore, we define $\mathsf{TF}(\mathbf{tr})$ as follows and present the details of two embedding layers subsequently:
\begin{align}
\label{eq:meta-transformer}
    \mathsf{TF}(\mathbf{tr}) & = \mathsf{TFEncoder}(\mathrm{Emb}(\mathcal{M})\oplus\mathrm{Emb}(\mathcal{E}), \mathcal{M}, \mathcal{E} \in \mathbf{tr}).
\end{align}
The above formulation is used for TGPT-2D without downstream features and TGPT-3D-\textbf{MTF} (detailed in Sec.~\ref{sec:fusion_omitting_feature}), as shown in Table~\ref{tab:trans-embed}. 

\paragraph{Metadata Embedding}
A transaction's metadata contains the basic information and key attributes commonly used in payment processing.
To make the foundation model general, we only include key metadata in model training, i.e., transaction time, amount, processing codes.
Though the location and merchant information are crucial metadata, we treat them as entities in $\mathcal{E}$ and encode them separately. Please refer to \textbf{Entity Embedding} for more details.

\begin{table*}[t]
    \centering
    \caption{The time-related fields in each transaction.}
    \centering
    \resizebox{0.65\textwidth}{!}{%
    \begin{tabular}{c|c}
        \toprule
        \textbf{Time Field} & \textbf{Explanation} \\
        \hline
        \texttt{time\_gap}  &  \begin{tabular}{@{}c@{}}\# days/hours between  the current \\ transaction and the previous transaction\end{tabular} \\
        \hline
        \texttt{month} & transaction month \\
        \hline
        \texttt{year} & \# years since 2000 \\
        \hline
        \texttt{day\_of\_week} & relative days in a week \\
        \hline
        \texttt{day\_of\_month} & relative days in the current month \\
        \bottomrule
    \end{tabular}
    }
    \label{tab:time-fields}
\end{table*}

Among the remaining metadata, temporal information is particularly critical.
The way time is encoded influences the model's ability to capture shopping behavior within each transaction sequence and the purchasing patterns over varying time spans.
Guided by time encoding approaches for time series~\cite{zerveas2021transformer} and point process~\cite{zhang2020self} and observed temporal patterns in consumer payments, we propose to convert each transaction's raw timestamp into the numerical fields listed in Table~\ref{tab:time-fields}.
Among these fields, \texttt{time\_gap} quantifies the interval between consecutive transactions, directly reflecting shopping cadence.
To capture seasonality and weekday/weekend effects, which are crucial for shopping activities, we further introduce four new fields to amplify these signals.
\texttt{month} and \texttt{year} encode the absolute calendar position, while \texttt{day\_of\_week} and \texttt{day\_of\_month} capture relative date information.
The unimodal and diverse time fields facilitate the shopping behavior modeling and simplify the time embedding together with other metadata.

After transforming timestamps into numerical values, all the metadata data types are either numerical or categorical.
We thereby design the metadata encoder as follows:
\begin{align}
\mathrm{Emb}(\mathcal{M}) = \concat_{c=0}^{C}\Bigl(\mathbf{E}_{m_c}\Bigr)\oplus\concat_{n=0}^{N}\Bigl(\mathsf{MLP}(log(m_n+\epsilon))\Bigr) ; m_c, m_n \in \mathcal{M}.
\end{align}
Specifically, we use a trainable embedding layer $\mathbf{E}_{m_c}\in \mathbb{R}^{|m_c|\times d_\mathcal{M}}$ to encode each categorical field $m_c \in \mathcal{M}$, where $|m_c|$ is the cardinality of $m_c$ and $d_\mathcal{M}$ is the embedding dimension for every metadata field in $\mathbf{tr}$.
For each numerical field $m_n \in \mathcal{M}$, we use an MLP to project it into $\mathbf{e}_{m_n}\in \mathbb{R}^{d_\mathcal{M}}$.
A log transformation is applied to $m_n+\epsilon$ to reduce data skewness, where $\epsilon$ is a small constant.
All encoded metadata are stacked ($\concat$ and $\oplus$) as a 2-D tensor with size $(C+N) \times d_\mathcal{M}$, where $C$ and $N$ are the number of categorical and numerical metadata fields, respectively.

\paragraph{Entity Embedding}
In a large-scale payment system, transactions can be grouped into segments based on their associated entities, as transactions tied to the same entity often exhibit pronounced similarities.
For example, transactions originating from a common geographic region tend to reflect local economic activity~\citep{aladangady2019transactions}.
Moreover, each merchant displays a distinctive transaction pattern and customer base.
Thus, a well-optimized and informative entity representation will further enhance the transaction representation learning.
We use a trainable embedding layer $\mathbf{E}_{e}$ to encode each entity $e \in \mathcal{E}$ into $d_{\mathcal{E}}$-dimensional embedding where $d_{\mathcal{E}} = d_{\mathcal{M}}$:
\begin{equation}
  \mathrm{Emb}(\mathcal{E}) =  \{\mathbf{E}_{e}\}, \space e \in \mathcal{E}.
\end{equation}

The primary challenge in jointly learning entity embedding with TGPT is handling the high-cardinality entities (e.g., merchant), which leads to a massive embedding table.
This significantly increases the model parameter size, thereby hindering its scalability and optimization.
Taking the merchant ID embedding table as an example, 10M merchants with embedding size 128 has 1.3B trainable parameters.
It will also yield a fat classification head when predicting future merchants, making the model optimization less efficient.

Motivated by the observation that merchants belonging to the same category or segments share similar properties and recent work in embedding table compression~\citep{shi2020compositional},
we employ compositional embedding to reduce the parameters of the merchant embedding layer.


\noindent \textbf{Compositional Embedding.}
For $n$ unique values of the entity $e \in \mathcal{E}$, we use the hashing trick to map each entity id to $k$ hashed values belonging to $(1,m), m \ll n$.
Then we initiate the embedding table  $\textbf{E}_e \in \mathbb{R}^{m \times d_{\mathcal{E}}}$, reducing the embedding table memory from $O(n \times d_{\mathcal{E}})$ to $O(m\times d_{\mathcal{E}})$.
Please check Appendix~\ref{appdx:comp_emb} for the complete implementation details.

\noindent \textbf{LLM-based Embedding Initialization.} Instead of learning entity embeddings from scratch, $\textbf{E}_e$ can also be initialized by embeddings pre-computed by another embedding model.
In this paper, we explore initializing MCC embeddings using LLMs and MCC descriptions~\citep{fan2025enhancing}.
Please refer to Appendix~\ref{appdx:llm_init} for technical details and Sec.~\ref{sec:exp-cls} for its contribution on the downstream task.

\noindent \textbf{Metadata Transformer Extended.} The Metadata Transformer can be extended to encode task-specific features, when available, alongside metadata fields and entities. In this case, Eq.~\ref{eq:meta-transformer} is modified as follows:
\begin{align}
\label{eq:meta-feat-transformer}
    \mathsf{TF}(\mathbf{tr}) & = \mathsf{TFEncoder}(\mathrm{Emb}(\mathcal{M})\oplus\mathrm{Emb}(\mathcal{E}) \oplus \mathrm{Emb}(\mathcal{F}), \mathcal{M}, \mathcal{E}, \mathcal{F} \in \mathbf{tr}).
\end{align}
This formulation is applied for TGPT-2D with downstream task features, as shown in Table~\ref{tab:trans-embed}.

\subsection{Modality Fusion}
\label{sec:modality_fusion}
As discussed in Sec.~\ref{sec:method-evolution}, integrating metadata and feature embeddings for each transaction presents two main challenges.
First, metadata embeddings are few but high-dimensional, while feature embeddings are numerous but low-dimensional, making direct integration difficult.
Second, after fusing these two modalities, the resulting transaction-level embedding needs to be sufficiently large to capture all relevant information; however, this significantly increases the computational cost of the temporal Transformer responsible for modeling inter-transaction dependencies.
We devise the following two approaches to handle the above challenges.

\subsubsection{Late Modality Fusion by Omitting Historical Feature Information}
\label{sec:fusion_omitting_feature}
This approach omits historical feature information for generative tasks and inputs only the concatenated metadata embeddings, processed by $\mathsf{TF}(\mathbf{tr})$ and of controllable size, into $\mathsf{TF}(\mathcal{S})$.
For downstream transaction classification, where feature information is essential, we fuse the future transaction embedding ${E}^{\mathbf{tr}_{n+1}}\in \mathbb{R}^{1 \times d_\mathbf{tr}}$ (output by $\mathsf{TF}(\mathcal{S})$) with the corresponding feature embeddings and pass them into $\mathsf{TF}(\mathcal{F})$.

Since the transaction embedding size $d_{\mathbf{tr}}$ is usually much larger than the feature embedding size $d_{f}$, we can rescale ${E}^{\mathbf{tr}_{n+1}}$ using an MLP to match $d_{f}$.
However, this approach constrains the information bandwidth, leading to information loss and suboptimal performance, as shown in Table~\ref{tab:txn_cls}. Instead, we divide ${E}^{\mathbf{tr}_{n+1}}$ evenly into multiple segments, each segment's size matches $d_{f}$.
These segments, together with all feature embeddings, are then input into the feature Transformer $\mathsf{TF}(\mathcal{F})$. The output embeddings from $\mathsf{TF}(\mathcal{F})$ are concatenated and connected to the prediction heads for downstream tasks.
Because this solution sequentially employs a metadata Transformer, a temporal Transformer, and a feature Transformer, we refer to it as TGPT-3D-\textbf{MTF}. The architecture is illustrated in Fig.~\ref{fig:main_figure}(d). For transactions $1$ to $n$, where features are not attached, the transaction embedding formulation follows Eq.~\ref{eq:meta-transformer}, as summarized in Table~\ref{tab:trans-embed}.

\begin{figure}
    \centering
    \includegraphics[width=0.7\linewidth]{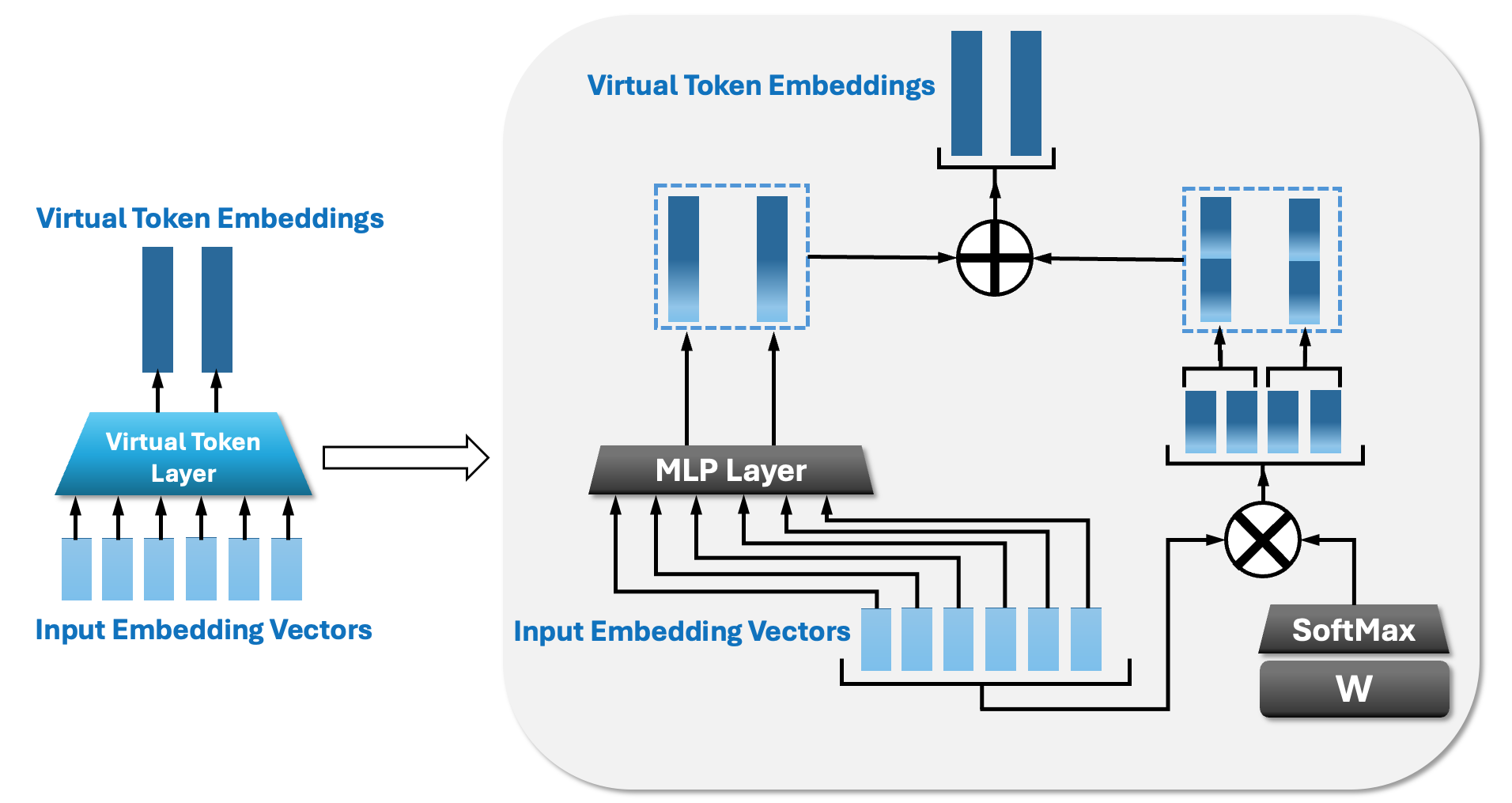}
    \caption{Virtual Token Layer (VTL).}
    \label{fig:virtual_token_layer}
\end{figure}

\subsubsection{Early Modality Fusion via Virtual Tokens}
\label{sec:virtual_token}
TGPT-3D-MTF addresses the aforementioned two challenges by omitting historical feature information.
However, this raises the question: is it possible to balance embedding dimension compatibility, information bandwidth, and computational feasibility while still retaining historical features, which may be valuable for both generative and downstream tasks?
To achieve this, we introduce the \textbf{V}irtual \textbf{T}oken \textbf{L}ayer (\textbf{$\mathsf{VTL}$}), which transforms a set of embeddings into an arbitrary number of “virtual tokens”.
These virtual tokens are soft representations that preserve essential information by providing sufficient bandwidth and flexible embedding sizes, and can be seamlessly passed to subsequent Transformer layers.
As illustrated in Fig.~\ref{fig:virtual_token_layer}, a $\mathsf{VTL}$ consists of both linear and nonlinear channels:

\noindent \textbf{Linear.}
A softmax-based linear combination enables the selection and aggregation of the most informative input embeddings.
Specifically, a parameter matrix $\textbf{W}$ is processed through a softmax layer to produce a coefficient matrix, which is then multiplied with the embedding matrix to generate $n$ new embeddings—each is a distinct linear combination of the originals.
This linear operation preserves maximum information from the original embeddings and supports effective gradient propagation. 
Since it does not alter the embedding dimensionality, larger virtual tokens can be formed by stacking multiple embeddings, while smaller tokens can be created by segmenting an embedding as needed.

\noindent \textbf{Nonlinear.}
A single-layer MLP with a nonlinear activation function is used to rescale the original embedding matrix into any desired number and size of output vectors.

The final virtual tokens are obtained via the summation of the linear and nonlinear components' outputs. 
In essence, the $\mathsf{VTL}$ functions as a scalable version of a ResNet~\citep{he2016deep} block: it enables arbitrary changes in token size, employs a linear path to retain original information, and uses a nonlinear path to enhance expressive power.
We utilize $\mathsf{VTL}$ in two ways within our architecture:

\noindent \textbf{Feature Tokenization.}
After the feature Transformer $\mathsf{TF}(\mathcal{F})$ processes the feature embeddings, a $\mathsf{VTL}$ is utilized to generate $v_f$ \textit{virtual feature tokens} ($v_f \ll |\mathcal{F}|$) with higher dimensionalities:
\begin{align}
    \mathbf{E}^{\mathcal{F}} = \mathsf{VTL}(\mathsf{TF}(\mathcal{F})) = \mathsf{VTL}(\mathsf{TFEncoder}(\mathcal{F})) = [({V}_{{1}}^\mathcal{F})^\top, ({V}_{{2}}^\mathcal{F})^\top, \ldots, ({V}_{{v_f}}^\mathcal{F})^\top]^\top,
\label{eq:virtual_feat_tokens}
\end{align}
where tensor $\mathbf{E}^{\mathcal{F}} \in \mathbb{R}^{v_f \times d_\mathcal{M}}$ and vector ${V}_{{i}}^\mathcal{F} \in \mathbb{R}^{1 \times d_\mathcal{M}}$ for $ 1\leq i \leq v_f$.
$\mathbf{E}^{\mathcal{F}}$ is then combined with the $\mathrm{Emb}(\mathcal{M})$ and jointly input to $\mathsf{TF}(\mathbf{tr})$.

\noindent \textbf{Transaction Tokenization.}
The virtual feature embeddings $\mathbf{E}^{\mathcal{F}}$ are stacked with metadata embeddings $\mathrm{Emb}(\mathcal{M})\oplus\mathrm{Emb}(\mathcal{E})$, fed into a Metadata Transformer encoder, and then passed through a second $\mathsf{VTL}$ to produce \textit{virtual transaction tokens}, which encapsulate all essential information for a single transaction. Thus, transaction representation $\mathbf{E}^{\mathbf{tr}}$ becomes a set of virtual token embeddings:
\begin{align}
\label{eq:virtual_trans_tokens}
    \mathbf{E}^{\mathbf{tr}} &= \mathsf{VTL}(\mathsf{TFEncoder}(\mathrm{Emb}(\mathcal{M})\oplus\mathrm{Emb}(\mathcal{E}) \oplus\mathsf{VTL}(\mathsf{TFEncoder}(\mathcal{F})))) \\\nonumber
    &= [({V}^{\mathbf{tr}}_{1})^\top, ({V}^{\mathbf{tr}}_{2})^\top, \ldots, ({V}^{\mathbf{tr}}_{v_t})^\top]^\top,
\end{align}
where tensor $\mathbf{E}^{\mathbf{tr}} \in \mathbb{R}^{v_t \times \frac{d_\mathbf{tr}}{v_t}}$ and vector ${V}_{{i}}^\mathbf{tr} \in \mathbb{R}^{1 \times \frac{d_\mathbf{tr}}{v_t}}$ for $ 1\leq i \leq v_t$. 
The $v_t$ virtual transaction tokens of all $n$ transactions are lined up sequentially, forming the $v_t n$ input vectors for the temporal Transformer $\mathsf{TF}(\mathcal{S})$. Consequently, with this dual application of $\mathsf{VTL}$, the architecture sequentially deploys a feature Transformer, a metadata Transformer, and a temporal Transformer; we refer to it as TGPT-3D-\textbf{FMT}, as illustrated in Fig.~\ref{fig:main_figure}(e).

Key advantages of this design include:
(1) Alignment of data modalities: virtual feature tokens can be sized to match $d_{\mathcal{M}}$, enabling effective fusion and joint processing by subsequent Transformers.
(2) Flexible bandwidth and efficiency: $v_f$ and $v_t$ can be tuned to ensure sufficient information bandwidth while keeping computational costs manageable.
(3) Optimized representation for sequential modeling: by representing each $\mathbf{tr}$ as a set of consecutive virtual tokens, the model avoids the prohibitive parameter and computational costs of large embedding sizes in $\mathsf{TF}(\mathcal{S})$.
Instead, it leverages Transformer architectures’ strengths in handling long sequences, efficiently encoding both complex intra- and inter-transaction dependencies.

Further discussion on our design of the virtual token mechanism and its connections to visual tokenization and dual-channel feature crossing methodologies are presented in Sec~\ref{sec:tokenization-feat-cross}. The formulation differences introduced by the virtual token mechanism for transaction-level embedding computation, compared with other TGPT variants, are summarized in Table~\ref{tab:trans-embed}.

\subsection{Time Complexity Analysis}
\label{sec:complexity}
The time complexity of the TGPT model is primarily driven by the self-attention operation performed across different modalities.
To simplify our analysis, we consider a single-head, single-layer self-attention architecture under context window $w$ with $\mathcal{F}$ available.
The complexities of the different TGPT variants are as follows:
\begin{itemize}[leftmargin=*]
\item \textbf{TGPT-1D:} This model exclusively uses the $\mathsf{TF}(\mathcal{S})$ module, as shown in Fig.~\ref{fig:main_figure}. Its time complexity is: $O_{\text{1D}} = O(wd_{\mathbf{tr}}^2 + w^2d_{\mathbf{tr}})$. This consists of the projection layer complexity ($O(wd_{\mathbf{tr}}^2)$) and the attention operation complexity ($O(w^2d_{\mathbf{tr}})$).
\item \textbf{TGPT-2D:} This model adds a $\mathsf{TF}(\mathbf{tr})$ to encode both merchant ($\mathcal{M}$) and feature ($\mathcal{F}$) modalities. The resulting complexity is: $ O_{\text{2D}}=O\left(\left((|\mathcal{M}|+|\mathcal{F}|)d_{\mathcal{M}}^2 + (|\mathcal{M}|+|\mathcal{F}|)^2d_{\mathcal{M}}\right)w\right) + O_{\text{1D}}$. 
\item \textbf{TGPT-3D-MTF:} This model only encodes $\mathbf{tr}_{n+1}$'s features. Its complexity includes the time complexities of $\mathsf{TF}(\mathbf{tr})$ for $\mathcal{M}$, $\mathsf{TF}(\mathcal{F})$ for $\mathbf{tr}_{n+1}$'s $\mathcal{F}$, and TGPT-1D: $O_{\text{3D-MTF}}=O\left(\left(|\mathcal{M}|d_{\mathcal{M}}^{2} + |\mathcal{M}|^{2}d_{\mathcal{M}}\right)w\right) + O\left(\left(|\mathcal{F}|+\frac{d_{\mathbf{tr}}}{d_f}\right)d_{f}^{2} + \left(|\mathcal{F}|+\frac{d_{\mathbf{tr}}}{d_f}\right)^{2} d_{f}\right) + O_{\text{1D}}$.
The term $\frac{d_{\mathbf{tr}}}{d_f}$ denotes the number of segments from the $E^{\mathbf{tr}_{n+1}}$ embedding fed into the $\mathsf{TF}(\mathcal{F})$.
\item \textbf{TGPT-3D-FMT:} This model applies the $\mathsf{VTL}$ for feature and transaction tokenization.
Its computation includes $\mathsf{TF}(\mathcal{F})$ for $w$ $\mathbf{tr}$s, $\mathsf{TF}(\mathbf{tr})$ for $|\mathcal{M} + v_f|$ tokens of $w$ $\mathbf{tr}$s, and $\mathsf{TF}(\mathcal{S})$ for $v_tw$ tokens with dimension $\frac{d_{\mathbf{tr}}}{v_t}$.
The total time complexity is: $O_{\text{3D-FMT}}=O\left(\left(|\mathcal{F}|d_{f}^2 + |\mathcal{F}|^{2}d_{f}\right)w\right) + O\left(\left((|\mathcal{M}| + v_f)d_{\mathcal{M}}^2 + (|\mathcal{M}| + v_f)^{2}d_{\mathcal{M}}\right)w\right) + O\left(v_tw(\frac{d_{\mathbf{tr}}}{v_t})^2+(v_tw)^2\frac{d_{\mathbf{tr}}}{v_t}\right)$. 
\end{itemize}

Although TGPT-2D is more expensive than TGPT-1D due to the Transformer encoding of metadata and features, and TGPT-3D-FMT is less efficient than TGPT-3D-MTF because it encodes every $\mathbf{tr}$'s feature,
we find that the added model capacity delivers substantial downstream gains (see Sec.~\ref{sec:exp-cls}).
Notably, FMT incurs only $\sim\times2.5$ higher inference latency than MTF while remaining within the company’s service-level agreement (SLA), preserving production viability.

We emphasize that novel information fusion techniques in TGPT-3D-FMT makes it significantly more efficient than TGPT-2D, as we have a large amount of transaction features requiring relatively small embedding sizes, and TGPT-3D-FMT's transaction embedding sizes are much greater than its attention window size:
\begin{theorem}
\label{thm_1}
        If $1 <d_f \ll d_{\mathcal{M}},$ $1<v_f, v_t, |\mathcal{M}|\ll |\mathcal{F}|$, and $1<w\ll \frac{d_{\mathbf{tr}}}{v_t}$,  then $O_{\mathrm{3D-FMT}} \ll O_{\mathrm{2D}}$. 
\end{theorem}
The proof of Theorem~\ref{thm_1} is in Appendix~\ref{appdx:proof}.

\subsection{Model Output and Loss Function}
\label{sec:loss_function}

As shown in Fig.~\ref{fig:main_figure},
for TGPT-1D and -2D, ${E}^{\mathbf{tr}_{n+1}}$ is connected to the prediction heads to generate metadata fields and downstream task labels.
For TGPT-3D-MTF, the output of $\mathsf{TF}(\mathcal{F})$ is only used for downstream task prediction, and the output of $\mathsf{TF}(\mathcal{S})$ is connected to metadata prediction heads.
For TGPT-3D-FMT, the last $k$ output vectors of $\mathsf{TF}(\mathcal{S})$, namely the Temporal Transformer encodings of the last k virtual tokens ${V}^{\mathbf{tr}}_{v_t-k}, {V}^{\mathbf{tr}}_{v_t-k+1}, \ldots, {V}^{\mathbf{tr}}_{v_t}$ of the $(n+1)$th transaction,  are concatenated and connected to the prediction heads.

TGPT has two types of optimization objectives:
the self-supervised objective for the next $\mathbf{tr}$ prediction (generative) and the supervised objective with downstream task labels (predictive).
The self-supervised training follows the heuristics that $\mathbf{tr_i}$ is dependent on $[\mathbf{tr_1}, \cdots,\mathbf{tr_{i-1}}]$ for $\mathbf{tr_i} \in \mathcal{S}$.
Optimizing TGPT via letting it predict key properties of future transactions based on historical transactions enables the model to learn the invariants behind transacting dynamics.
Meanwhile, we believe the joint optimization of self-supervised and supervised signals can mutually benefit both the generative and predictive tasks.

To make self-supervised learning generalize to all transactions and downstream tasks, we pick \textit{four} essential fields from $\mathcal{M}$ to be generated for future transaction: \textit{time\_gap}, \textit{amount}, \textit{merchant}, and \textit{MCC}.
time\_gap uses its original integer value (numerical); transaction amount is an integer representing US dollar cents (numerical);
each merchant/MCC has a unique identifier (categorical).
The prediction head and loss function for self-supervised training are designed according to the field type we would like to predict:
\begin{align}
\label{eq:sst_loss}
\mathcal{L}_{\mathrm{sst}} (num) & = (\mathsf{MLP}({E}^{\mathbf{tr}_{n+1}}) - \log (y_{num} + \epsilon))^{2},
\\
\mathcal{L}_{\mathrm{sst}} (cat) & = - \sum^{C}_{c=1}y_c\log (softmax(\mathsf{MLP}({E}^{\mathbf{tr}_{n+1}}))_c).
\end{align}

\noindent \textbf{Weight Tying.}
Since the prediction head for high-cardinality categorical entity (e.g., merchant) introduces a massive amount of trainable parameters, we resort to weight tying technique~\cite{press2017using} to reduce model size and improve training/inference efficiency.
Specifically, instead of using a dedicated prediction head $\mathrm{MLP}$, the entity logits is computed via $\mathbf{E}_e ({E}^{\mathbf{tr}_{n+1}})^\top$, where $\mathbf{E}_e$ is the trainable embedding layer parameters of entity $e$ from $\mathrm{Emb}(\mathcal{E})$. 

In TGPT, the supervised signal comes from downstream tasks inside the company which is usually a classification problem.
The final loss function is the composition of self-supervised loss and an optional supervised loss based on the availability of downstream tasks, plus regularization losses: weight decay and the $\mathsf{VTL}$ matrix orthogonality losses (Sec~\ref{sec:virtual_token}):
\begin{align}
\mathcal{L} = \lambda_1 \mathcal{L}_{\mathrm{sst}} (num) + \lambda_2\mathcal{L}_{\mathrm{sst}} (cat) + \lambda_3\mathcal{L}_{\mathrm{st}} + \lambda_4\mathcal{L}_\mathrm{weight} + \lambda_5(\mathcal{L}_\mathrm{orth}^{\mathcal{F}} + \mathcal{L}_\mathrm{orth}^{\mathbf{tr}}), 
\end{align}
where the $\lambda$s are the weight hyper-parameters; $\mathcal{L}_{\mathrm{st}}$ is dependent on the availability of downstream task supervision signal; $\mathcal{L}_W$ stands for the weight decay loss.
$\mathsf{VTL}$ matrix orthogonality losses for virtual feature tokens and virtual transaction tokens are given by
\begin{align}
    \mathcal{L}_\mathrm{orth}^{\mathcal{F}} & = || \mathbf{E}^{\mathcal{F}}_\mathrm{lin} (\mathbf{E}^{\mathcal{F}}_\mathrm{lin})^\top - \mathbf{I}_{v_f}||^2_\mathrm{F}, \\
    \mathcal{L}_\mathrm{orth}^{\mathbf{tr}} & = || \mathbf{E}^{\mathbf{tr}}_\mathrm{lin} (\mathbf{E}^{\mathbf{tr}}_\mathrm{lin})^\top - \mathbf{I}_{v_t}||^2_\mathrm{F}, 
\end{align}
where $|| \cdot ||^2_\mathrm{F}$ denotes Frobenius norm, and $\mathbf{E}^{\mathcal{F}}_\mathrm{lin}$ and $\mathbf{E}^{\mathbf{tr}}_\mathrm{lin}$ respectively denote the linear combination part of the feature and transaction virtual tokens. 

In Sec.~\ref{sec:train_tips}, we have additional discussion on model design and lessons learn through model training.

%% file: sec04-recipe.tex
\section{Training Recipe}
\label{sec:method-train-receipe}


\subsection{Data Preparation}
\label{sec:exp-setup}
We collect and process billion-scale transactions in strict compliance with data security and privacy regulations, ensuring no access or disclosure of personally identifiable information.
We segment a full transaction sequence into length-$l$ subsequences, padding those shorter than $l$ and discarding those below $l_{\min}$.
To have a fair comparison with baselines,
we curate three datasets for three evaluation tasks shown in Table~\ref{tab:dataset}, then train and evaluate TGPT on each of them separately.
The test data consists of transaction sequences collected from a period subsequent to the training data.

\noindent \textbf{T-JGC:} it is our primary evaluation setting based on a key revenue-driven product.
It is used to demonstrate TGPT's transaction generation and anomaly detection performance against a strong production model.
T-JGC data merges mainstream (metadata only) with downstream transactions (metadata + features).
The merger yields a richer payment history but introduces rows that contain metadata only (Fig.~\ref{fig:mmtt_data}), increasing modeling complexity.

\noindent \textbf{T-RES:} it evaluates TGPT's dining trajectory generation performance and merchant embedding quality using side information.
T-RES data uses offline, card-present restaurant transactions only with metadata.
We filter out sequences where the same merchant appears in $>80\%$ of transactions to filter na\"ive recurring patterns and prevent model overfitting.

\noindent  \textbf{T-MCC:} to have a fair comparison of TGPT and LLM-based foundation model, we benchmark them under the future transaction MCC prediction task.
T-MCC data is curated by sampling 1M transaction sequences with diverse types of MCCs ($\sim800$ classes), which is appropriate for LLM fine-tuning.

\subsection{Model Config and Training Stats}
Due to the large design space of TGPT and its dependence on proprietary data, infrastructure capacity, and downstream tasks, there is no single silver-bullet model configuration.
Instead, we report the search space of key model components explored during tuning in Table~\ref{tab:hyperparameter}, along with model sizes of the most performant configurations trained on the T\_JGC dataset: with the embedding layer and prediction heads excluded, TGPT-
1D and TGPT-2D contain 7.5M parameters, while TGPT-3D-FMT contains 10.3M parameters.

We use the AdamW optimizer, CosineAnnealingLR scheduler, and gradient clipping during training; weight decay and dropout are enabled with fixed default values.
The batch size is empirically determined based on VRAM capacity and training stability.
An extended discussion of the design choices we made and the lessons learned during TGPT training can be found in Sec.~\ref{sec:train_tips}.

All models are trained using PyTorch 2.1 with DistributedDataParallel on GPU clusters equipped with NVIDIA A100 80GB GPUs and CUDA 12.3.
The most performant TGPT-1D, 2D, and 3D configurations require 362, 712, and 480 GPU-hours, respectively, to converge on the T\_JGC dataset.

%% file: sec05-exp.tex
\section{Model Evaluation}
We evaluate TGPT's foundational capability through three diverse tasks towards different payment use cases: \textbf{T-JGC}: future transaction generation jointly with downstream anomaly transaction detection (Sec.~\ref{sec:exp-cls});
\textbf{T-RES}: dining transaction trajectory generation and merchant representation learning (Sec.~\ref{sec:exp-gen});
\textbf{T-MCC}: benchmarking MCC prediction performance with fine-tuned LLMs (Sec.~\ref{sec:exp-pred}).
Sec.~\ref{sec:exp-efficiency} shows the empirical impact of several scalability designs in TGPT.
Our findings and insights through experiments are highlighted in \textit{italics}.

\begin{table}[]
    \centering
    \caption{Dataset statistics.}
    \resizebox{0.85\textwidth}{!}{%
    \begin{tabular}{l|c|c|c|c|c|c}
    \toprule
        \textbf{Dataset} & \textbf{Seq\_len} & \textbf{$\#$ Train Seqs} & \textbf{$\#$ Test Seqs} & \textbf{$\#$Merchants} & \textbf{$\#$Classes} & \textbf{$|\mathcal{F}|$}\\
        \hline
        T-JGC & 8 &  100M & 10M & 3M & 2 & $\sim400$ \\
        \hline
        T-RES &  16 & 200M & 20M & 500K & 500K & -\\ 
        \hline
        T-MCC &  16 & 1M  & 100K & 50K  &  $\sim800$ & - \\ 
    \bottomrule
    \end{tabular}
    }
    \label{tab:dataset}
\end{table}

\subsection{Transaction Generation and Anomaly Detection}
\label{sec:exp-cls}

\textbf{Evaluation Setting.} With the downstream feature available, we assess TGPT's capacity to learn comprehensive payment dynamics through generating future transaction metadata and its anomaly probability (Table~\ref{tab:txn_cls}).
For MCC and merchant, the average Recall@1 is reported.
It is particularly challenging for merchant prediction, as TGPT must identify the correct one from 3M distinct options.
MAE and MSE are reported for time and amount prediction.
A proprietary metric (\textbf{AD}) is used to assess \textbf{A}nomaly \textbf{D}etection performance, which measures the accuracy under an operational threshold.
We omit the detailed formulation of \textbf{AD} and only report the relative improvement (\textbf{RI}) of each model w.r.t. the production model for company policy compliance.
Importantly, the production model is trained on the downstream dataset without missing fields, while TGPT is trained on the merged dataset containing bubbles.

We compare TGPT variants along with two representative baselines in this experiment.
For transaction generation, a Transformer-based sequential recommendation model, \textbf{Transformers4Rec}~\cite{moreira2021transformer}, is trained to generate future MCCs.
To investigate the contribution of sequential information to anomaly detection performance, we train an instance-level (per transaction) \textbf{Feat-Transformer} that only processes a single transaction without temporal context.
To evaluate the contribution of individual architectural components of TGPT, we compare the performances of the following model variants:

\begin{itemize}[leftmargin=*]
    \item \textbf{TGPT-1D}: with/without downstream features (Fig.~\ref{fig:main_figure}(b)).
    \item \textbf{TGPT-2D}: with/without downstream features (Fig.~\ref{fig:main_figure}(c)).
    \item \textbf{TGPT-3D-MTF} (Fig.~\ref{fig:main_figure}(d)): with segments or MLP-Scaling.
    \item \textbf{TGPT-3D-FMT} (Fig.~\ref{fig:main_figure}(e)): 
    \begin{itemize}[leftmargin=*]
        \item \textbf{FMVTL}: with Feature and Metadata VTLs; the last $k=1$ virtual temporal tokens are connected to prediction heads (unless otherwise specified, $k=1$ is by default).
        \item \textbf{FMVTL-2}: same with \textbf{FMVTL} except $k=2$.    
        \item \textbf{FMVTL-nonlin}: Feature and Metadata VTLs only adopts the nonlinear (MLP) component. 
        \item \textbf{FMVTL-lin-map}: Feature and Metadata VTLs are replaced by simple linear mappings.
        \item \textbf{FVTL}: with Feature VTL but without Metadata VTL. 
        \item \textbf{FMVTL+LLM}: The MCC embeddings are initialized using LLM-generated representations, where ISO MCC descriptions serve as input prompts.
    \end{itemize}    
\end{itemize}

\noindent \textbf{Transaction Generation.} Table~\ref{tab:txn_cls} shows that TGPT-1D/2D w/o Feat achieves reasonable performance on metadata prediction, correctly identifying MCCs (out of $\sim800$ categories) with near 50\% accuracy and merchants (out of 3M) with $>30\%$ accuracy.
Transformers4Rec yields notably lower accuracy than TGPT‑1D/2D for future MCC generation.
As the merchant ID possesses much higher cardinality than MCC and is therefore more difficult to model, merchant prediction result for Transformers4Rec is omitted.
Across amount and time prediction tasks, the 2D variants consistently achieve lower prediction errors than TGPT‑1D, suggesting that the additional Metadata Transformer enables the modeling of more fine‑grained information, while delivering comparable performance on MCC and merchant prediction.
We further observe that feature‑free models perform on par with their feature‑augmented counterparts across most metrics, suggesting that auxiliary feature information is not fully exploited in generative tasks.
TGPT-3D-MTF variants show no obvious advantage over TGPT-1D/2D w/ Feat in most cases, indicating that the MTF approach to feature fusion is suboptimal for generative tasks.
TGPT‑3D‑FMT achieves the best performance across all metrics, underscoring the critical role of feature integration and a well‑designed modality fusion architecture.

\noindent \textbf{\textit{Remarks}.} \textit{ 1. Domain-engineered features provide substantial enhancement to the generative task.  2. Effective multimodal integration is non-trivial: features maximize prediction performance the most when they are properly fused with metadata before temporal modeling.
3. FMT with dual VTLs successfully leverages the predictive signals of domain features when they are appropriately integrated.} 

\begin{table}[t]
     \caption{Performance comparison of TGPT and its variants on transaction generation and anomaly detection (\textbf{AD}) tasks. 
     For compliance, only the relative improvement (\textbf{RI}) over the production model performance is shown for the anomaly detection task.
     }
     \centering
    \resizebox{0.95\textwidth}{!}{%
    \begin{tabular}{l|c|cc|cc|cc}
        \toprule
        \textbf{Model}  & \textbf{AD (\%)}  & \textbf{MCC  (\%)} & \textbf{Mrch  (\%)} & \multicolumn{2}{c|}{\textbf{Amount }} & \multicolumn{2}{c}{\textbf{Time}} \\
        \hline
          & \textbf{RI$\uparrow$}  & \textbf{Rec@1$\uparrow$} & \textbf{Rec@1$\uparrow$} & \textbf{MAE$\downarrow$} & \textbf{MSE$\downarrow$} & \textbf{MAE$\downarrow$} & \textbf{MSE$\downarrow$} \\
        \hline
        \textbf{Transformers4Rec} & -  & 33.46 & - & - & - & - & - \\
        \hline
        \textbf{Feat-Transformer} & +7.7  & - & - & - & - & - & - \\
        \hline
        \textbf{TGPT-1D}  &   &  &  &  &  \\
        \quad - w/o Feat & -90.5  & 48.86 & 31.60  & 1.29 & 3.84 & 0.1350 & 0.0355 \\
        \quad - w/ Feat & -9.1  & 49.25 & 30.46 & 1.30 & 3.93 & 0.1370 & 0.0282 \\
        \hline
        \textbf{TGPT-2D} &   &  &  &  &  \\
        \quad - w/o Feat  & -87.0   & 48.89 & 31.41 & 1.27 &  3.78  & \underline{0.0702} & 0.0100  \\
        \quad - w/ Feat & +7.3  & 49.17 & 30.38 & 1.26 & 3.79 & 0.0740 & \underline{0.0095} \\       
        \hline
        \textbf{TGPT-3D-MTF}  &   &  &  &  &  &  &  \\
         \quad - Segments & +14.6  & 48.23 & 32.41 & 1.33 & 4.05 & 0.1420 & 0.0329  \\
         \quad - MLP-Scaling & +9.5  & 48.17 & 32.29 &  1.32 & 4.05   & 0.1400  & 0.0325   \\
        \hline
        \textbf{TGPT-3D-FMT} &    &  &  &  &  \\
        \quad - FMVTL  &  \underline{+19.2}  & 49.94 & 32.08  & \textbf{1.23} & \textbf{3.71}  & 0.0893 & 0.0120 \\
        \quad - FMVTL-2  & +15.5 &  \textbf{50.12} & \underline{32.70} & \underline{1.24} & \underline{3.73} & 0.0800 & 0.0098 \\
        \hline
        \quad - FMVTL-nonlin & +6.7 & 49.55 & 31.20 & 1.30 & 3.86 & 0.0901 & 0.0141 \\
        \quad - FMVTL-lin-map & +0.2 & 49.62 & 31.06 & 1.26 & 3.78 & 0.0880 & 0.0152 \\
        \quad - FVTL & -12.9 & 48.45 & 30.91 & 1.28  & 4.16 & 0.0767 & 0.0111 \\ 
        \hline
        \quad - FMVTL + LLM & \textbf{+22.5} & \underline{50.01} & \textbf{32.73}  & 1.25 & 3.77  & \textbf{0.0686} & \textbf{0.0072} \\
        \bottomrule
    \end{tabular}
    }
    \label{tab:txn_cls}
\end{table}

\noindent \textbf{Anomaly Detection.} Domain features play a more critical role in this task:
TGPT-1D w/ Feat exhibits degraded performance on metric \textbf{AD} compared to both the production and Feat-Transformer baselines. In contrast, the 2D variants and Feat‑Transformer explicitly use transformers to model interactions among features, enabling them to outperform the 1D architecture.
Surprisingly, TGPT-2D w/ Feat shows no gains over the Feat-Transformer, despite utilizing full sequences versus single transactions, highlighting the challenge of effectively utilizing temporal information.
Sec.~\ref{sec:train_tips} presents further analysis from an information bandwidth perspective. 
The MTF fusion strategy in TGPT-3D-MTF provides considerable gains, achieving substantial relative improvement over the baselines, proving successful at integrating sequential metadata with features.
However, this architecture is constrained by its inability to incorporate feature information from historical transactions. 
Finally, TGPT-3D-FMT emerges as the optimal configuration, delivering the most significant performance improvements on the anomaly detection metric while simultaneously maintaining superior generative performances. 

\noindent \textbf{\textit{Remarks}.}\textit{ 1. Naive feature concatenation for each transaction (2D) fails to leverage sequential and domain feature information for the classification objective. 2. Sophisticated modality fusion architectures (3D-MTF/FMT) are more critical for downstream classification than for the generative task. 3. MTF/FMT is unaffected by data bubbles. 4. The dual VTL structure in FMT enables more effective multi-modal representation learning, successfully bridging the gap between generation and classification objectives. }

\noindent \textbf{Ablation Study.}
We systematically evaluate the effectiveness of key architectural components for modality fusion.
TGPT-3D-MTF: Segmented transaction embeddings maintain complete information bandwidth between the transaction embedding and the feature Transformer, outperforming MLP-scaling on metric $\textbf{AD}$, confirming that information compression degrades downstream performance. 
TGPT-3D-FMT: FMVTL-nonlin eliminates the linear combination in VTL, converting it to a standard MLP.
Universal performance drops indicate that the linear pathway is essential for gradient flow and information preservation; FMVTL-lin-map replaces the full VTL with basic linear projection for dimension matching.
Further degradation on $\textbf{AD}$ demonstrates that nonlinear expressiveness is crucial, validating the linear+nonlinear design.
FVTL removes metadata VTL entirely, causing the most severe performance drops across most classification and generation metrics. 
This confirms that both FM and FMT fusion stages are necessary for effective multi-modal integration. 

FMVTL+LLM yields modest improvements on generative metrics and delivers notable gains over FMVTL trained from scratch on the anomaly detection task, demonstrating that LLM-derived semantic information enhances downstream performance by leveraging external knowledge to better understand transaction dynamics.

\noindent \textbf{LLM-initialized MCC Embedding.}
FMVTL+LLM yields modest improvements on generative metrics and delivers notable gains over FMVTL trained from scratch on the classification task, demonstrating that LLM-derived semantic information enhances downstream performance by leveraging pre-existing knowledge to better understand transaction dynamics.

\begin{figure}
    \centering
\includegraphics[width=0.98\linewidth]{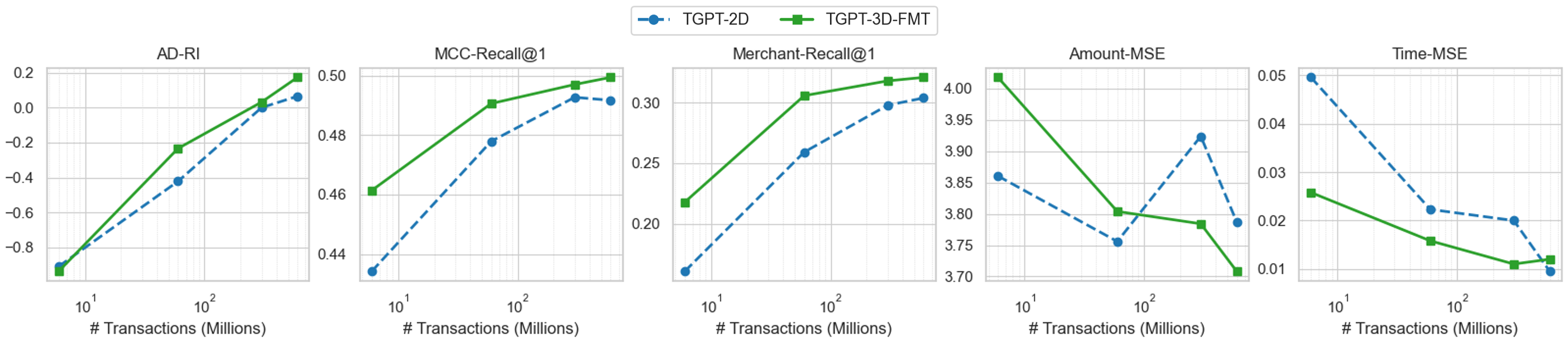}
    \caption{Data scaling behavior: performance metrics as a function of the number of training transactions. The x-axis is in log scale.}
    \label{fig:data_scaling}
\end{figure}

\noindent \textbf{Data Scaling Law.} We investigate model performance under increasing training transaction volumes by training multiple instances of TGPT‑2D/3D.
As shown in Fig.~\ref{fig:data_scaling}, TGPT's performance improves consistently with larger training datasets across all tasks, indicating that the data scaling law is one key driver of further improving the generative and predictive power of transaction foundation models. Notably, the downstream task metric \textbf{AD} exhibits the greatest performance gains under the 3D configuration. On the smallest dataset, however, the 3D model performs comparably to the 2D model, indicating that the advantages of the 3D architecture emerge more prominently with larger data volumes. Given the modest size of our current models, there remains substantial room for scaling model capacity, which we leave for future work.

\begin{table}
    \centering
    \caption{(a) Future restaurant and (b) future location prediction accuracy of TGPT-2D given previous dining trajectories.}
    \begin{minipage}{0.62\textwidth}
        \centering
        \resizebox{\textwidth}{!}{%
        \begin{tabular}{l|cccccc}
        \toprule
             \makecell[c]{\textbf{(a)}}& \textbf{SASRec} & \textbf{TGPT-1D} & \textbf{TGPT-2D} & \makecell[c]{\textbf{TGPT-2D} \\ \textbf{-m-only}} \\
            \hline
            \textbf{Rec@1}  & 11.9\% & 12.8\% & \textbf{14.2\%} & \underline{13.7\%} \\
            \textbf{Rec@50}  & 38.5\% & 42.5\% & \textbf{45.6\%} & \underline{43.8\%} \\
        \bottomrule
        \end{tabular}
        }
        \label{tab:restaurant_gen}
    \end{minipage}%
    \hfill
    \begin{minipage}{0.33\textwidth}
        \centering
        \resizebox{\textwidth}{!}{%
        \begin{tabular}{l|ccc}
        \toprule
             \makecell[c]{\textbf{(b)}} & \makecell[c]{\textbf{Zip} \\ \textbf{code}} & \textbf{City} & \textbf{State} \\
            \hline
            \textbf{Top-1} & 28\% & 42\% & 84\% \\ 
            \textbf{Top-10} & 60\% & 69\% & 91\% \\ 
        \bottomrule
        \end{tabular}
        }
        \label{tab:exp-loc}
    \end{minipage}
\end{table}

\subsection{Dining Trajectory Generation}
\label{sec:exp-gen}

We evaluate the dining trajectory generation capability of TGPT via training TGPT on a restaurant transaction only data (T-RES) and letting TGPT predict the next restaurant.
Since there is no personal identifiable information in the input data and our data and model weights are secured internally, the dining trajectory generation task only reflects the foundation transaction modeling capability of TGPT from a collective perspective without compromising individual privacy.

\noindent \textbf{Quantitative Evaluation.}
We select SASRec~\citep{kang2018self}, a classic sequential recommendation model trained on merchant IDs, as the restaurant prediction baseline.
Since there is no downstream features in this data, we only evaluate the TGPT-1D and -2D models.

Table~\ref{tab:restaurant_gen}(a) shows the restaurant prediction performance under Recall@1 and Recall@50.
The results show that all TGPT model variants outperform SASRec.
Notably, among 45.6\% test sequences, TGPT-2D can rank the exact future restaurant as the top 50 candidates from 500K restaurants.  
Meanwhile, TGPT-2D outperforms TGPT-1D, demonstrating metadata Transformer encoder is better than na\"ive MLP-based embedding layer.
TGPT-2D optimized with all prediction heads is better than TGPT-2D only optimized with the merchant prediction head (TGPT-2D-m-only).
\textit{The results collectively verify that the MMTT foundation model's multi-field optimization is advantageous and necessary: predicting auxiliary metadata fields improves merchant prediction.}

Thanks to the additional restaurant location information we have, we also evaluate the future dining location prediction accuracy of TGPT-2D using restaurant prediction as a proxy.
Specifically, there is no location info in TGPT input/output, and the accuracy is computed by checking if the location of generated top-k restaurants match the location of the ground truth restaurants.
The evaluation results in Table~\ref{tab:exp-loc}(b) are very promising: we can successfully predict the future dining locations with Accuracy@10=69\% even if TGPT does not ingest any location-related info. 
The results further suggest that \textit{TGPT is able to encode consumer payment patterns and merchant similarities and thereby benefit dining trajectory generation.}

\begin{figure}
    \centering
\includegraphics[width=0.98\linewidth]{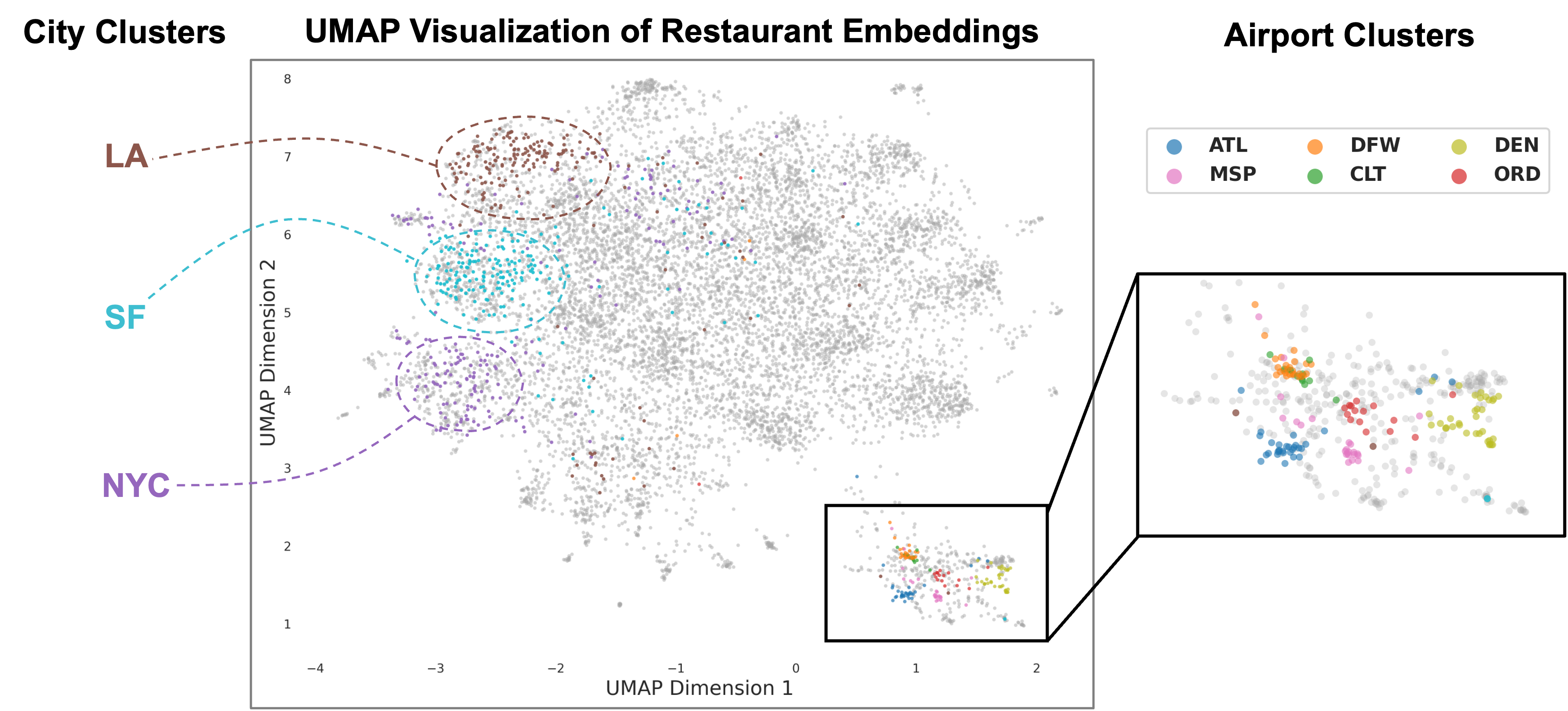}
    \caption{UMAP visualization of the TGPT merchant embeddings of the top 10K restaurants ranked by transaction volume, plus highlighted restaurants from major US cities (left) and airports (right).}
    \label{fig:res-vis}
\end{figure}

\noindent \textbf{Qualitative Evaluation.}
The stellar quantitative results on dining trajectory generation suggest that TGPT learns the inherent similarity between restaurants.
To validate it, we visualize the restaurant embeddings learned from TGPT via UMAP~\citep{mcinnes2018umap} in Fig.~\ref{fig:res-vis}.
Based on the heuristic that the restaurants belonging to the same dining trajectory should have location proximity with each other, we highlight restaurants from selected US cities and airports to demonstrate the proximity from a local mobility and air mobility perspective, respectively.

From the local mobility perspective, we highlight 200 randomly sampled restaurants from each of three major US cities: Los Angeles (LA), San Francisco (SF), and New York City (NYC).
In Fig.~\ref{fig:res-vis} (left), most restaurants belonging to the same city are closely clustered together, and three city-based clusters have explicit boundaries between each other.
It proves that TGPT embeddings learned by predicting future transactions encode the location proximity of different restaurants very well.

From the air mobility perspective (Fig.~\ref{fig:res-vis} (right)), we highlight restaurants from the top 2 hub airports ranked by passenger numbers from 3 major US airlines (Delta: ATL and MSP; American: DFW and CLT; United: DEN and ORD).
The UMAP plot shows that airport-based restaurants are densely clustered and clearly separated with other restaurants, suggesting the airport-based restaurants have their unique properties and are different from most day-to-day dining places.

We have the following three observations from the individual airport view:
(1) Restaurants located within the same airport tend to form distinct clusters, and for hub airports operated by the same airline, these clusters appear relatively close to each other.
(2) The embeddings of DFW (orange) and CLT (green) restaurants show substantial overlap, which could be due to the high passenger traffic and shared hub status of these two largest American Airlines hubs.
(3) ORD (Chicago) restaurant embeddings are positioned between those of DEN, DFW, and MSP.
This intermediate positioning may reflect ORD’s role as a major hub for both United and American Airlines, as well as potential similarities in passenger demographics with nearby airports such as MSP.

\textit{To summarize, the UMAP visualization explains restaurant embeddings learned by TGPT do encode the proximity of restaurants along the dining trajectories. }


\subsection{Merchant Category Prediction}
\label{sec:exp-pred}
Besides the dining trajectory generation (i.e., merchant prediction) task, we evaluate the transaction generation performance on another representative transaction field: merchant category code (MCC).
The curated T-MCC data includes transactions covering $\sim800$ MCCs.
Instead of comparing classic sequential prediction baselines, we implement LLM-based baselines, as LLM-based foundation models are actively explored by many industry practitioners.

Since this task does not have downstream features, we compare TGPT-2D's next MCC prediction performance with representative open weights LLMs~\citep{touvron2023llama, jiang2023mistral7b, javaheripi2023phi}.
For LLM baselines, we describe each historical transaction's metadata using plain English language following a table-to-text template~\citep{hegselmann2023tabllm} and use it as the LLM input. LLM is fine-tuned to predict the next transaction's MCC, where each MCC is registered as a new special token in each LLM's tokenizer.
Please refer to Appendix~\ref{appdx:llm_pred} for more LLM baseline implementation details.

\begin{figure}
    \centering
\includegraphics[width=0.9\linewidth]{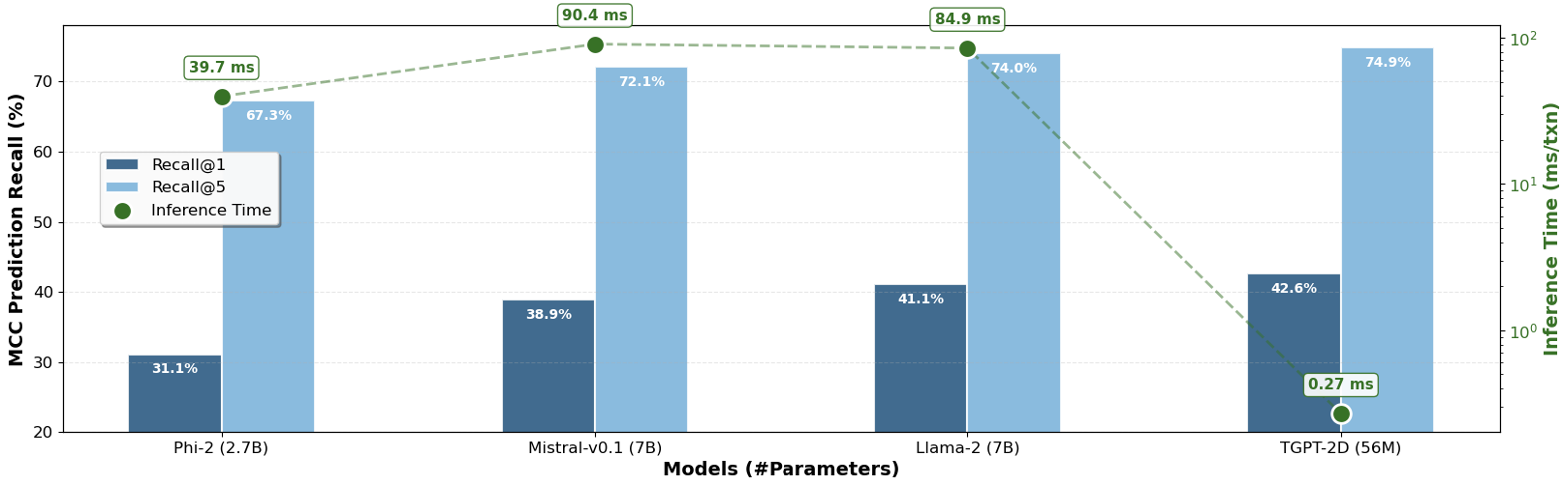}
    \caption{The field prediction task performance and inference speed comparison between TGPT and other models.}
    \label{fig:field_pred}
\end{figure}

In Fig.~\ref{fig:field_pred}, we report the MCC prediction performance (blue bars) of different models and their respective inference time (green dots) on a single NVIDIA A100 GPU (80GB) instance.
For a fair comparison, we incorporate the embedding layer size when computing TGPT-2D's model size.
TGPT-2D obtains better prediction performance than the best LLM
(Llama2-7B) with 92\% fewer parameters (56M vs. 7B) and
300× faster inference speed (0.27 ms vs. 84.9 ms).
It demonstrates that \textit{foundational models tailored to MMTT data is cost-efficient and practical comparing to LLM-based solutions on the MCC prediction task}.
Meanwhile, we are impressed by the MCC prediction performance of fine-tuned LLMs and are optimistic about LLM-based foundation models on suitable transaction downstream tasks.

\subsection{Scalability Evaluation}
\label{sec:exp-efficiency}


In Sec.~\ref{sec:method}, we devise and employ various techniques to improve TGPT's scalability.
Table~\ref{tab:efficiency} shows their relative impact on peak VRAM usage and anomaly detection performance w.r.t. the TGPT-3D-FMT base model without any scalability design.
Since the uncompressed high-cardinality merchant embedding table takes numerous parameters, the compositional embedding and weight tying substantially reduce TGPT model size and thereby peak memory usage.
Meanwhile, we observe that tying the softmax weight with the embedding table only has a minor negative influence on downstream performance ($-$0.6\%), proving that weight tying is a practical solution.
More surprisingly, the compositional embedding and local attention reduce the embedding parameters and context window but yield slightly better performance on the downstream task.
We speculate that \textit{compact model alleviates the overfitting, and the selected anomaly detection task may not favor long-term transaction dependencies.}

\begin{table}
    \centering
      \caption{The relative single-GPU peak VRAM usage and anomaly detection performance of different scalability enhancing techniques w.r.t the base TGPT-3D model.}
    \resizebox{0.75\textwidth}{!}{%
    \begin{tabular}{l|c|c}
    \toprule
        \textbf{Scalability Design}  & \textbf{Peak Memory Usage} &
        \makecell[c]{\textbf{Downstream} \\ \textbf{Performance}} \\
        \hline
        TGPT-3D-base  & - & -\\
        + compositional embedding (Sec.~\ref{sec:method-txn-encoder})  & -22.6\% &  +1.9\%\\
        + local attention (Sec.~\ref{sec:method-seq-encoder})  & -24.9\% &   +2.8\%\\
        + weight tying (Sec.~\ref{sec:loss_function}) & -73.8\% &  -0.6\% \\
    \bottomrule
    \end{tabular}
    }
    \label{tab:efficiency}
\end{table}


%% file: sec06-discuss.tex
\section{Discussion on Model Design and Training}
\label{sec:train_tips}
We have been carrying out numerous experiments with different TGPT model variants and training approaches.
Due to privacy and legal compliance, we cannot disclose full details of the model architecture and source code.
We distill and share the following practical guidelines aimed at improving training stability, task performance, and computational efficiency. To improve readability, we present the material following the model’s forward propagation—i.e., in the order that data flows through each component.

\noindent \textbf{Normalizations.} Applying normalization when connecting embeddings to transformer modules is essential for both training stability and performance.
For TGPT, BatchNorm consistently outperformed LayerNorm, likely due to the highly diverse cardinalities across fields and the multi‑modal nature of the data.
Within each transformer, we adopt a Pre-LN architecture as recommended by~\citep{xiong2020transformer}.  


\noindent \textbf{Positional Encoding for Non‑sequential Fields.} For $\mathsf{TF}(\mathbf{tr})$ and $\mathsf{TF}(\mathcal{F})$, where input fields lack intrinsic order, we compared one-hot and sinusoidal positional encodings.
One‑hot encodings delivered slightly better performance.


\noindent \textbf{Information Bandwidth.}
In the \textbf{TGPT-2D} configuration, the output embedding size $d_{\mathbf{tr}}$ of the \emph{Metadata Transformer} $\mathsf{TF}(\mathbf{tr})$ determines its communication bandwidth with the subsequent \emph{Temporal Transformer} $\mathsf{TF}(\mathcal{S})$, as denoted by TGPT-2D-$d_{\mathbf{tr}}$ in Fig.~\ref{fig:data_scaling}. Increasing $d_{\mathbf{tr}}$ from $64$ to $128$ yields substantial performance improvements---particularly for metrics A and B---underscoring the importance of high-bandwidth information transfer. However, because the key-field embeddings produced by $\mathsf{TF}(\mathbf{tr})$ are concatenated to form the full transaction-level embedding that serves as input to $\mathsf{TF}(\mathcal{S})$, enlarging $d_{\mathbf{tr}}$ proportionally increases the embedding dimensionality and, consequently, the computational cost of $\mathsf{TF}(\mathcal{S})$. Scaling $s_{\mathbf{tr}}$ further from $128$ to $256$ results in OOM errors on an NVIDIA~A100 GPU ($80\,$GB). 

\noindent \textbf{Integrating Metadata Transformer Outputs.}
As detailed in \textbf{Information Bandwidth}, constructing a transaction‑level embedding from the $\mathsf{TF}(\mathbf{tr})$ output for input into the $\mathsf{TF}(\mathcal{S})$ poses a significant design challenge in TGPT‑2D. We evaluate five integration strategies (design choices for the $\mathsf{Integrate}$ operation in Table~\ref{tab:trans-embed}) and observe their respective trade‑offs: 
\begin{itemize}[leftmargin=*]
    \item Concatenation. Directly concatenating all output embeddings from the $\mathsf{TF}(\mathbf{tr})$ to form the transaction‑level embedding offers maximum information bandwidth, but results in prohibitively large embeddings and frequent out‑of‑memory (OOM) errors. 
    \item MLP compression / linear mapping. All output embeddings are concatenated and then projected to an arbitrary dimensionality using an MLP block (with linear or nonlinear activation). This allows flexible bandwidth control, but the large input dimensionality produces dense parameter matrices prone to training instability and overfitting, leading to sub‑optimal performance.
    \item Pooling (summation/average/max). Element‑wise pooling is applied across all outputs of the $\mathsf{TF}(\mathbf{tr})$. This approach is computationally efficient and memory‑friendly, but aggregates hundreds of heterogeneous field embeddings without filtering, introducing substantial noise. The resulting transaction‑level embedding has the same dimensionality as a single field embedding, severely limiting information bandwidth.
    \item Field-embedding selection + pooling. To reduce noise, only embeddings from a small set of essential fields are pooled. While this improves signal quality, the dimensionality remains constrained to that of a single field embedding, limiting bandwidth. Additionally, field selection risks information loss: it is difficult to identify the most critical fields among hundreds, and pooling inherently discards information.
    \item Field-embedding selection + concatenation. This approach replaces pooling with concatenation for the selected key‑field embeddings, reducing noise and offering more flexible bandwidth. While pooling‑related information loss is avoided, selection still risks discarding relevant features and requires strong heuristics, often informed by domain expertise.
\end{itemize}
These trade‑offs motivate the design of \textbf{TGPT-3D-FMT}, which incorporates the \emph{virtual token mechanism} to decouple information bandwidth from embedding size, thereby providing full controllability in information flow between transformers of different modalities.
Among the above methods, the option of field-embedding selection + concatenation provided a relatively good balance of stability, efficiency, and accuracy when combined with carefully selected fields based on domain knowledge and extensive experimentation. 
Table~\ref{tab:metadata_integration} in Appendix~\ref{appdx:metatransformer_integration} provides an empirical comparative summary of the strengths and limitations associated with the five integration strategies.

\noindent \textbf{Design Choices for Temporal Transformer.}
We investigated the encoder-only Transformer architecture for $\mathsf{TF}(\mathcal{S})$ with bi-directional attention.
During training, the last $j$ transactions of each sequence are masked for prediction, where $1 < j < |\mathcal{S}|$, as the BERT-style~\citep{devlin2019bert}.
For single-step prediction ($j=1$), the encoder-only model's performance on generating transaction attributes (merchant, MCC, time, amount) is comparable to a decoder-only model.
To generate multiple future transactions, the encoder-only model operates recursively, predicting the next $j$ transactions while discarding the oldest $j$ to maintain a fixed sequence length $|\mathcal{S}|$.

Empirically, this recursive approach allows the encoder-only model to outperform the decoder-only model for short-term forecasting ($1 < j < |\mathcal{S}|$).
However, its primary weakness is context erosion: after generating $|\mathcal{S}| - 1$ transactions, the original input is completely lost, causing subsequent predictions to lack grounding.
In contrast, the decoder-only model's auto-regressive nature retains the full historical input, making it superior for long-horizon forecasting ($j > |\mathcal{S}|$).
Therefore, the optimal Transformer architecture depends on the prediction horizon required by the specific use case.

\noindent \textbf{Anomaly Prediction Loss.} Given that anomalies are rare events, the downstream classification labels are highly imbalanced.
We tested focal loss~\citep{lin2018focalloss} and class weighting strategies, but found no improvement over standard cross‑entropy loss with equal class weights.
We attribute this to the large absolute number of anomaly samples in our dataset, which mitigates the impact of class imbalance despite its rarity.

\noindent \textbf{Parameter Precision and Quantization.} We experimented with BF16 and FP32 precision during training.
BF16 helps constrain overfitting on small datasets but limits performance growth when scaling data volume.
In contrast, FP32 tends to overfit small datasets but better preserves scaling gains with larger data.

These lessons directly informed the architectural and optimization choices in TGPT‑3D, improving stability, generalization, and scalability across diverse data regimes.


%% file: sec07-rw.tex
\section{Related Work}

\subsection{Deep Learning for Transaction Data}
Most of the transaction data has rich entity-connectivity and temporal information, which make it suitable for graph or sequential neural encoders.
The graph-based methods regard the transactions between different entities as edges in a graph (e.g, account-to-account, account-to-merchant).
Graph Neural network (GNN)-based~\citep{liu2021intention, li2023diga} approaches are used to learn transaction or other entity representations.
The graph models capture contextual information in transaction networks and are ideal for tasks like item recommendation and anomaly detection with transaction data.
Sequential methods encode a series of transactions along the temporal dimension and leverage RNN~\citep{babaev2019rnn, skalski2023towards} or Transformer~\citep{padhi2021tabular, zhang2023fata, unbox2025behaviorgpt, li2025panther, aminian2025fraudtransformer, li2025panther} to encode the transaction sequences.
The sequential models are either trained in a supervised fashion~\citep{babaev2019rnn, aminian2025fraudtransformer} to directly accommodate downstream tasks or together with self-supervised objectives that reconstruct or generate certain transaction properties~\citep{padhi2021tabular, bazarova2024universal, unbox2025behaviorgpt, li2025panther}.
The common downstream tasks for sequential modeling of transaction data include classification and regression.

In this work, we deal with with consumer payment sequences with diverse downstream tasks.
Therefore, we adopt the sequential modeling paradigm and train the model with self-supervised and supervised objectives.
We highlight the differences between our work and previous work with sequential transaction encoders~\citep{babaev2019rnn, skalski2023towards, padhi2021tabular, zhang2023fata, unbox2025behaviorgpt, aminian2025fraudtransformer, li2025panther} as follows:
1) TGPT is trained on billions of real-world consumer-to-merchant transaction data and evaluated on different downstream tasks against production models;
2) instead of using discrete tokens to model transaction property and sequence, TGPT leverages a hybrid approach to encode MMTT data toward optimal information bandwidth between different data dimensions;
3) we explore how TGPT integrates LLM knowledge and compare TGPT with LLM-based baselines.

\subsection{Transformer for Tabular and Temporal Data}
Our transaction data's MMTT format resembles the tabular, time series, and marked point process data.
Many recent works have leveraged the Transformer to model the above data types.
We review the representative works that motivated TGPT and compare their differences with our work.

TabNet~\citep{arik2021tabnet} and TabTransformer~\citep{huang2020tabtransformer} motivate our metadata and feature Transformers design, especially the numerical and categorical field encoding.
The tabular foundation models~\citep{somepalli2021saint, he2023anameta, hollmann2025accurate, jingangtabicl} also inspire us to devise a foundation model for MMTT data.
In contrast to most tabular foundation models, which only work on small-scale and low-dimensional tabular data, TGPT is validated on a more complex and large-scale MMTT data with multiple data modalities.
Unlike recent foundation models targeting large-scale relational data~\citep{feykumorfm, ranjan2025relational}, TGPT is designed around the sequential structure of payment transactions and tailored to Visa’s downstream tasks.

Most research on time series Transformers focus on better encoding of timestamps~\citep{zerveas2021transformer, zhou2021informer, garza2023timegpt, liuitransformer, ekambaram2024tiny}.
Those works motivate us in devising time embedding, temporal Transformer, and the self-supervised objective for time information.
However, MMTT data is more complex than time series data: 1) MMTT has high-dimensional heterogeneous data fields for each timestamp, while time series data only has unimodal data type; 2) MMTT has irregular intervals between timestamps, while time series data has uniform intervals.
Therefore, the above time series Transformers cannot be directly adopted in MMTT data.

Comparing to the time series data, the transaction sequence is more like a marked point process~\citep{mei2017neural} composed of discrete events with irregular intervals.
Previous work use Transformer to model the point process's intensity function~\citep{zuo2020transformer, zhang2020self, yang2021transformer} or the distribution of events and inter-event times~\citep{panos2024decomposable}.
Due to the multi-modal nature and multi-scale time info of the MMTT data, modeling it as a point process with Transformers will result in a fragmented model with excessive amount of parameters for each yet to be defined event type.
Therefore, we resort to LLM's next token prediction paradigm and using MSE loss for time prediction toward a unified and coherent Transformer architecture in TGPT.

Recent endeavors leverage Transformer-based LLMs to encode tabular~\citep{ruan2024language} and temporal~\citep{zhang2024large} data by translating the data into text.
In this work, we do not adopt the above approach because:
1) predictive modeling does not require textual output, which is different from tabular QA and tabular understanding tasks with rich semantics;
2) MMTT data has high-dimensional metadata and features, and employing LLM to encode them is costly;
3) recent research~\citep{tan2024language, van2024position} corroborates that LLMs do not have a significant advantage on many temporal forecasting and tabular classification tasks.

\subsection{Continuous Space Tokenization and Feature Crossing}
\label{sec:tokenization-feat-cross}
As detailed in the paper, the goal of designing the virtual token mechanism is to map continuous latent embeddings into a set of discrete tokens, enabling effective integration with another modality (such as metadata) or along another data dimension (such as temporal) while allowing flexible information bandwidth. A similar objective exists in other domains, such as computer vision and time series modeling. However, our realization of this objective employs a methodology resembling techniques developed for feature crossing.

\noindent \textbf{Continuous Space Tokenization}. Vision Transformer (ViT)~\citep{dosovitskiy2020image} splits an image into fixed-size patches, linearly projects each patch to an embedding, adds position encodings, and feeds the sequence to a Transformer. More recently, in vision–language models (VLMs)~\citep{alayrac2022flamingo, liu2023visual, zhou2024transfusion}, continuous latent embeddings from the image modality are tokenized into multiple patch embeddings and then fused into sequential language models. For tokenizing continuous spaces, pre-trained vector-quantized variational autoencoder (VQ-VAE) techniques are commonly applied~\citep{van2017neural,razavi2019generating,esser2021taming}. Time-LLM~\citep{jin2023time,jin2024position}, as a representative Foundational time-series model~\citep{liang2024foundation}, segments sequences into consecutive patches for tokenization and reprograms them with a small set of text prototypes that semantically describe each patch’s basic properties, resembling the role of the discrete codebook in VQ-VAE from another approach. In transaction modeling, however, pre-trained vector-quantization models are not readily available. Moreover, the latent representations of metadata, features, or transactions generally lack explicit physical semantics, unlike numerical time series that can be naturally described with terms such as ``up'' or ``down''. Consequently, our approach aligns more closely with the ViT-style methodology, augmented with feature crossing.

\noindent \textbf{Feature Crossing.} Based on our extensive experience in transaction and feature modeling, feature crossing and interaction-based techniques have proven effective for extracting information from tabular data that contains diverse metadata fields and large numbers of numerical features~\citep{xu2024transnet}. Hence, instead of applying a simple linear projection, as adopted in ViT, in our approach, the $\mathsf{VTL}$ architecture employs a dual-channel design reminiscent of Wide \& Deep~\citep{cheng2016wide}, DCN~\citep{wang2017deep}, and our prior TransNet design~\citep{xu2024transnet}. Unlike DCN, however, $\mathsf{VTL}$ does not explicitly perform complex feature crossing, as both preceding and subsequent Transformer blocks implicitly model these interactions. Our ablation study demonstrates the performance advantage of the dual-channel feature crossing design over the ViT-style linear projection (FMVTL-lin-map in Table~\ref{tab:txn_cls}).

%% file: appendix.tex
\section*{Appendix}

\section{Compositional Embedding}
\label{appdx:comp_emb}

Algorithm~\ref{fig:comp_embed} shows the pseudocode of the compositional embedding class.
In class initialization, a trainable EmbeddingBag~\footnote{https://docs.pytorch.org/docs/stable/generated/torch.nn.EmbeddingBag.html} layer is created to store $m+1$ embeddings with dimension $=dim$ (Lines 2-5).
$k$ hash functions (e.g., MurmurHash3~\footnote{https://scikit-learn.org/stable/modules/generated/sklearn.utils.murmurhash3\_32.html}) with different random seeds map the total number of entities $n$ to a smaller shared embedding space of size $m$.
The hashed indices are stored as non-trainable parameters in a lookup table, and the index 0 is reserved for unseen entities (Lines 6-14).
During the forward pass (Lines 16-20), the input entity index $e$ is mapped through the lookup table to obtain its hashed indices (Line 17), and the corresponding embeddings of hashed indices are aggregated using EmbeddingBag with mean pooling to obtain the output compositional embedding (Line 18).


        
        
        
    

\begin{algorithm}
\begin{algorithmic}[1]

\Procedure{Init}{$n, m, dim, k$}
    \State Initialize \textbf{EmbeddingBag} layer:
    \State \hspace{1em} Number of embeddings $\gets m + 1$
    \State \hspace{1em} Embedding dimension $\gets dim$
    \State \hspace{1em} Padding index $\gets 0$
    \State Initialize empty list \textit{hashed\_idx}
    \For{$i \gets 0$ to $k-1$}
        \State $idx \in \mathbb{R}^{n+1} \gets$ array of integers from $0$ to $n$
        \State $idx \gets$ \textit{murmurhash3\_32}($idx$, seed $= i$)
        \State $idx \gets$ ($idx \bmod m$) $+ 1$
        \State Set $idx[0] \gets 0$ \Comment{Padding index}
        \State Append $idx$ to \textit{hashed\_idx}
    \EndFor
    \State $lookup\_table \in \mathbb{R}^{k\times (n+1)} \gets$ stack arrays in \textit{hashed\_idx}
\EndProcedure

\Procedure{Forward}{$e$}
    \State $e\_indices \in \mathbb{R}^{k} \gets$ \textit{lookup\_table} indexed by $e$
    \State $e\_embedding \in \mathbb{R}^{m} \gets$ \textbf{EmbeddingBag}($e\_indices$)
    \State \Return $e\_embedding$
\EndProcedure

\end{algorithmic}
\caption{Compositional Embedding Class.}
\label{fig:comp_embed}
\end{algorithm}

\section{Proof of Theorem 1}
\label{appdx:proof}
\textbf{Theorem 1.} \textit{If $1 <d_f \ll d_{\mathcal{M}},$ $1<v_f, v_t, |\mathcal{M}|\ll |\mathcal{F}|$, and $1<w\ll \frac{d_{\mathbf{tr}}}{v_t}$,  then $O_{\mathrm{3D-FMT}} \ll O_{\mathrm{2D}}$. }
\begin{proof}
Substituting the time complexities in $O_{\mathrm{3D-FMT}} \ll O_{\mathrm{2D}}$ by their definitions, we need to show: 
\begin{multline*} \left(|\mathcal{F}|d_{f}^2 + |\mathcal{F}|^{2}d_{f}\right)w + \left((|\mathcal{M}| + v_f)d_{\mathcal{M}}^2 + (|\mathcal{M}| + v_f)^{2}d_{\mathcal{M}}\right)w + v_tw\left(\frac{d_{\mathbf{tr}}}{v_t}\right)^2+(v_tw)^2\frac{d_{\mathbf{tr}}}{v_t} \\ \ll \left((|\mathcal{M}|+|\mathcal{F}|)d_{\mathcal{M}}^2 + (|\mathcal{M}|+|\mathcal{F}|)^2d_{\mathcal{M}}\right)w + wd_{\mathbf{tr}}^2+w^2d_{\mathbf{tr}}
\end{multline*}
We analyze the inequality by comparing the terms on the left-hand side (LHS) with those on the right-hand side (RHS). 

\noindent\textbf{Comparison of terms related to $\mathsf{TF}(\mathbf{tr})$ and $\mathsf{TF}(\mathcal{F})$:}
\begin{itemize}[leftmargin=*]
\item LHS terms: $\left(|\mathcal{F}|d_{f}^2 + |\mathcal{F}|^{2}d_{f}\right)w + \left((|\mathcal{M}| + v_f)d_{\mathcal{M}}^2 + (|\mathcal{M}| + v_f)^{2}d_{\mathcal{M}}\right)w$
\item RHS terms: $\left((|\mathcal{M}|+|\mathcal{F}|)d_{\mathcal{M}}^2 + (|\mathcal{M}|+|\mathcal{F}|)^2d_{\mathcal{M}}\right)w$ \end{itemize}
Given the conditions $|\mathcal{M}| \ll |\mathcal{F}|$ and $v_f \ll |\mathcal{F}|$, we have the asymptotic approximations $(|\mathcal{M}|+|\mathcal{F}|) \approx |\mathcal{F}|$ and $(|\mathcal{M}| + v_f) \ll |\mathcal{F}|$.
The dominant part of the RHS is therefore approximately $|\mathcal{F}|d_{\mathcal{M}}^2 + |\mathcal{F}|^2d_{\mathcal{M}}$. Now we show that each component of the LHS is asymptotically smaller than the corresponding dominant component of the RHS: \begin{itemize}[leftmargin=*]
\item Since $d_f \ll d_{\mathcal{M}}$, we have $|\mathcal{F}|d_{f}^2 \ll |\mathcal{F}|d_{\mathcal{M}}^2$ and $|\mathcal{F}|^2d_f \ll |\mathcal{F}|^2d_{\mathcal{M}}$.
\item Since $(|\mathcal{M}| + v_f) \ll |\mathcal{F}|$, we have $(|\mathcal{M}| + v_f)d_{\mathcal{M}}^2 \ll |\mathcal{F}|d_{\mathcal{M}}^2$ and $(|\mathcal{M}| + v_f)^2d_{\mathcal{M}} \ll |\mathcal{F}|^2d_{\mathcal{M}}$.
\end{itemize}
As every term on the LHS is asymptotically negligible compared to a corresponding term on the RHS, their sum is also asymptotically negligible.

\noindent \textbf{Comparison of terms related to $\mathsf{TF}(\mathcal{S})$:}
\begin{itemize}[leftmargin=*]
\item LHS terms: $\alpha_1 = v_tw\left(\frac{d_{\mathbf{tr}}}{v_t}\right)^2+(v_tw)^2\frac{d_{\mathbf{tr}}}{v_t} = \frac{w d_{\mathbf{tr}}^2}{v_t} + v_t w^2 d_{\mathbf{tr}}$
\item RHS terms: $\alpha_2 = wd_{\mathbf{tr}}^2+w^2d_{\mathbf{tr}}$
\end{itemize}
To prove $\alpha_1 \ll \alpha_2$, we analyze their ratio $R = \frac{\alpha_1}{\alpha_2}$: \[ R = \frac{\frac{w d_{\mathbf{tr}}^2}{v_t} + v_t w^2 d_{\mathbf{tr}}}{wd_{\mathbf{tr}}^2+w^2d_{\mathbf{tr}}} = \frac{d_{\mathbf{tr}}^2/v_t + v_t w d_{\mathbf{tr}}}{d_{\mathbf{tr}}^2+w d_{\mathbf{tr}}} = \frac{d_{\mathbf{tr}}/v_t + v_t w}{d_{\mathbf{tr}}+w} \] 
Let's factor out $d_{\mathbf{tr}}$ from the numerator and denominator: \[ R = \frac{\frac{d_{\mathbf{tr}}}{v_t}\left(1 + \frac{v_t^2 w}{d_{\mathbf{tr}}}\right)}{d_{\mathbf{tr}}\left(1 + \frac{w}{d_{\mathbf{tr}}}\right)} = \frac{1}{v_t} \cdot \frac{1 + \frac{v_t^2 w}{d_{\mathbf{tr}}}}{1 + \frac{w}{d_{\mathbf{tr}}}} \] As $w \ll \frac{d_{\mathbf{tr}}}{v_t}$, both $\frac{v_t^2 w}{d_{\mathbf{tr}}}$ and $\frac{w}{d_{\mathbf{tr}}}$ approach 0.
Therefore, the ratio $R$ asymptotically approaches $\frac{1}{v_t}$: \[ \lim R = \frac{1}{v_t} \]
Given the condition $v_t > 1$, we have $1/v_t < 1$.
This shows that $\alpha_1$ is asymptotically a fraction of $\alpha_2$, representing a significant reduction.
Since the time complexity terms on the LHS are asymptotically smaller than their counterparts on the RHS, the sum on the LHS is asymptotically smaller than the sum on the RHS.
Thus, we have shown that $O_{\mathrm{3D-FMT}} \ll O_{\mathrm{2D}}$.
\end{proof}

\section{LLM-based Entity Embedding Initialization}
\label{appdx:llm_init}
Among all entities in the consumer payment ecosystem, merchant category code (MCC) is particularly semantic rich thanks to the ISO 18245~\footnote{https://www.iso.org/standard/79450.html} defining the meaning and description of each MCC code.
According to the Visa Merchant Data Standards Manual~\citep{visa2025manual}, we use the extended description of an MCC to prompt LLMs to generate the MCC embeddings.
Motivated by the one-word prompt design for sentence embedding~\citep{jiang2024scaling}, we design the following prompt template:
\begin{tcolorbox}[colback=blue!5!white, colframe=blue!75!black]
The MCC \textbf{\{mcc\}}, titled \textbf{\{title\}}, is described as follows: \textbf{\{description\}}. This category includes \textbf{\{included\}}, similar MCCs are \textbf{\{similar\}}. Please provide the embedding of the MCC \textbf{\{mcc\}} in the next token.
\end{tcolorbox}
\noindent The \{included\} is an extended list of merchant types belonging to the corresponding MCC.
The \{similar\} is a list of MCCs similar to the current MCC.
Those fields can be found in the Visa Merchant Data Standards Manual mentioned above.

By prompting the open-weights LLMs using the above prompt template, we extract the activations of an LLM's last layer as the embedding of the input MCC.
As the last layer's activations have included the context information and summarized the MCC information.
This solution is cost effective and scalable since no fine-tuning is required, and it is faster than sentence embedding models with bi-directional attentions, which makes it is possible to scale to high-cardinality entities.
Its LLM-agnostic design also enables us to adopt the latest open-weights LLMs and LLMs with various sizes.
Please find more details of applying sentence embeddings to foundation models in our extended work~\citep{fan2025enhancing}.

\section{LLM for Transaction Field Prediction}
\label{appdx:llm_pred}
In Sec.~\ref{sec:exp-pred} and Fig.~\ref{fig:field_pred}, we compare TGPT performance on next transaction MCC prediction against LLM-based methods.
We present the technical details of LLM-based methods, which can be generalized to predicting any field of future transactions.

Assuming the future transaction is dependent on historical transactions, we design the following text template for LLM fine-tuning:
\begin{tcolorbox}[colback=blue!5!white, colframe=blue!75!black]
[More historical transaction info...] \\
Target field: \textbf{\{field\_value\}}, transaction amount: \textbf{\{amount\}}, transaction date: \textbf{\{date\}}, $\cdots$, transaction time: \textbf{\{time\}}. \hfill $\triangleright$  \space $\texttt{last transaction info}$. \\
Target field: [MASK]. \hfill $\triangleright$ \space $\texttt{field to be predicted}$.
\end{tcolorbox}
\noindent Specifically, the descriptions of the key metadata of each historical transaction are used as the context.
One can flexibly adjust the number historical transactions and which transaction information to be described in the context. 
The target field value to be predicted for the future transaction is the last token of the text template, which is masked during training.

In the MCC prediction task, since the target field value (i.e., four-digit MCC codes) are not in the vocabulary of open-weights LLMs, we insert 800 new special tokens to each LLM's vocabulary to represent MCC codes.
Consequently, fine-tuning LLMs on this data is required to let LLMs understand transaction patterns and learn the correlation between historical transaction information and future field value represented by special tokens.

Although expanding a pretrained LLM's vocabulary will add overhead during fine-tuning, this approach is common in multi-modal LLM~\citep{zhan2024anygpt} and generative recommendation~\citep{geng2022recommendation}, and it facilitates model evaluation.
In contrast, using each MCC's plain-text name has following challenges: 1) each MCC's name includes 1 to 30 tokens (e.g., \textbf{MCC 4011}: Railroads and \textbf{MCC 4814}: Telecommunication Services, including Local and Long-Distance Calls, Credit Card Calls, Calls Through Use of Magnetic Stripe-Reading Telephones, and Fax Services); 2) some MCCs share similar words in their names (e.g., \textbf{MCC 5532}: Automotive Tire Stores and \textbf{MCC 5533}: Automotive Parts and Accessories Stores).
It is non-trivial to extract the exact MCC entity from a sequence of generated tokens.

In a nutshell, the LLM's fine-tuning and inference process is a next token prediction task with only one forward pass (i.e., only predicting the next one token).
In the experiment, we fine-tune three open-weights LLMs (Llama2-7B~\citep{touvron2023llama}, Mistral-7B-v0.1~\citep{jiang2023mistral7b}, Phi2-2.7B~\citep{javaheripi2023phi}) with different sizes and from different model families.
LoRA~\citep{hu2022lora} and FSDP~\citep{zhao2023pytorchfsdp} are used to accelerate fine-tuning and improve scalability.


\section{Practical Guide on Metadata Transformer Output Integration Strategies}
\label{appdx:metatransformer_integration}
Table~\ref{tab:metadata_integration} provides an empirical comparative summary of the strengths and limitations associated with the five integration strategies discussed in Sec~\ref{sec:train_tips}: concatenation, MLP compression (or linear mapping),  fooling (summation/average/max), field-embedding selection + pooling, and field-embedding selection + concatenation.

\begin{table}[]
    \centering 
    \caption{Empirical comparison of Metadata Transformer output integration strategies in TGPT-2D. The method adopted for results in Table~\ref{tab:txn_cls} is highlighted.} \label{tab:metadata_integration} 
    \small 
    \resizebox{0.995\textwidth}{!}{%
    \begin{tabular} {l p{3cm} p{3.5cm} p{1.5cm} p{3.5cm} p{4cm}}  
        \toprule 
        \textbf{Method} & \textbf{Bandwidth} & \textbf{Noise Level} & \textbf{OOM Risk} & \textbf{Information Loss} & \textbf{Performance} \\
        \midrule Concat & High & High (no filtering) & High & Low &  Impractical due to memory limits \\
        MLP compression & Flexible (adjustable via projection) & Moderate & Moderate & Moderate (compression removes detail) & Prone to overfitting; sub-optimal in practice \\
        Pooling (sum/avg/max) & Low (same as single field embedding) & High (aggregates all fields) & Low & High (pooling discards information) & Weak — signal diluted and bandwidth-limited \\
        Field selection + pooling & Low (same as single field embedding) & Lower than full pooling & Low & High (field omission + pooling) & Better than full pooling, but bandwidth-limited\\
        {Field selection + concat} & {Moderate–high (depends on \#fields)} & {Low (noise filtered)} & {Moderate} & {Moderate (field omission)} & Balance of stability, efficiency, and performance \\
        \bottomrule 
    \end{tabular} 
    }
\end{table}

\begin{table}
    \centering
    \caption{The search space of key model components.} \label{tab:hyperparameter} 
    \resizebox{0.6\textwidth}{!}{%
    \begin{tabular}{l c} 
        \toprule 
        \textbf{Component} & \textbf{Search Space}  \\
        \midrule \# Transformer layers &  [1,2,3,4] \\
        \# Attention heads &  [2,4,6,8]\\
        Metadata embedding dimension &  [4,8,16,32,64,128,256] \\
        Feature embedding dimension & [4,8,16,32,64,128] \\
        \# Virtual tokens & [1,2,4,8,16,32] \\
        \bottomrule 
    \end{tabular} 
    }
\end{table}

